\documentclass[journal,compsoc,twoside]{IEEEtran}

\usepackage{
	amsmath,
	amssymb,
	booktabs,
	footnote,
	graphicx,
	multirow,
	pgfplots,
	url,
    hyperref,
    siunitx,
    nicefrac,
    pifont,
    threeparttable,
    schemabloc,
    soul,
    verbatim,
    todonotes,
    float,
    textcomp
}
\usepackage[shortcuts]{extdash}
\usepackage[inline]{enumitem}

\usepackage[labelformat=simple]{subcaption}

\usepackage{tikz}
\usetikzlibrary{arrows,shapes,circuits.logic.US,shapes.gates.logic.IEC,calc,decorations.markings,patterns}
\tikzstyle{branch}=[fill,shape=circle,minimum size=3pt,inner sep=0pt]

\usepgfplotslibrary{groupplots}

\usepackage[graphicx]{realboxes}
\usepackage{stackengine}[2013-09-11]

\begin{document}

	\title{LUTNet: Learning FPGA Configurations for Highly Efficient Neural Network Inference}

	\author{
		Erwei~Wang,~\IEEEmembership{Student~Member,~IEEE},
		James~J.~Davis,~\IEEEmembership{Member,~IEEE},
		Peter~Y.~K.~Cheung,~\IEEEmembership{Senior~Member,~IEEE},
		and~George~A.~Constantinides,~\IEEEmembership{Senior~Member,~IEEE}%
		\IEEEcompsocitemizethanks{
			\IEEEcompsocthanksitem The authors are with the Department of Electrical and Electronic Engineering, Imperial College London, London, SW7 2AZ, United Kingdom.
			E-mail: \texttt{\{erwei.wang13, james.davis, p.cheung, g.constantinides\}@imperial.ac.uk}.
			\IEEEcompsocthanksitem The authors are grateful for the support of the United Kingdom EPSRC (grant number EP/P010040/1), Imagination Technologies, and the Royal Academy of Engineering.
		    \IEEEcompsocthanksitem Supporting data for this article are available online at \texttt{https://doi.org/10.5281/zenodo.3517151}.
		}
	}
	
	\markboth{IEEE Transactions on Computers}{Wang \MakeLowercase{\emph{et al.}}: LUTNet: Learning FPGA Configurations for Highly Efficient Neural Network Inference}

    \IEEEtitleabstractindextext{
    	\begin{abstract}
    	
    		Research has shown that deep neural networks contain significant redundancy, and thus that high classification accuracy can be achieved even when weights and activations are quantized down to binary values.
    		Network binarization on FPGAs greatly increases area efficiency by replacing resource-hungry multipliers with lightweight XNOR gates.
    		However, an FPGA's fundamental building block, the $K$-LUT, is capable of implementing far more than an XNOR: it can perform any $K$-input Boolean operation.
    		Inspired by this observation, we propose LUTNet, an end-to-end hardware-software framework for the construction of area-efficient FPGA-based neural network accelerators using the native LUTs as inference operators.
    		We describe the realization of both unrolled and tiled LUTNet architectures, with the latter facilitating smaller, less power-hungry deployment over the former while sacrificing area and energy efficiency along with throughput.
    		For both varieties, we demonstrate that the exploitation of LUT flexibility allows for far heavier pruning than possible in prior works, resulting in significant area savings while achieving comparable accuracy.
    		Against the state-of-the-art binarized neural network implementation, we achieve up to twice the area efficiency for several standard network models when inferencing popular datasets.
    		We also demonstrate that even greater energy efficiency improvements are obtainable.
    
    	\end{abstract}
    	
    }
    
    \maketitle

	\IEEEraisesectionheading{
	    \section{Introduction and Motivation}
	    \label{sec:intro}
	}
	
	    \IEEEPARstart{T}{hroughout} the last decade, general-purpose processors represented the dominant hardware platforms for deep neural network (DNN) inference.
	    However, able to attain significantly higher throughput and energy efficiency than their general-purpose counterparts, custom hardware implementations are becoming increasingly popular alternatives.
	    Field-programmable gate arrays (FPGAs) are attractive platforms for the realization of DNN inference applications due to their cost for low- to medium-volume deployment and facilitation of rapid time to market.
	
		During inference, the most common---and expensive---computational node in a DNN performs a function of the form in \eqref{eq:synapse_normal}, calculating a channel output $y$.
		Each weight $w_n$ is a constant determined during training, $\boldsymbol{x}$ a vector of $N$ channel inputs, and $f$ an activation function such as the widely used rectified linear unit.
		In the extreme case where $\boldsymbol{w} \in \left\{-1,1\right\}^{N}$---so-called binarized neural networks (BNNs)---the multiplications become cheap or free to implement.
		With weight inputs left variable, multipliers become XNOR gates.
		When networks are unrolled~\cite{Parhi}, weights are fixed, and so the XNOR gates can be further simplified into buffers and inverters, all of which are usually subsumed into the downstream adder logic.
		Also beneficial for BNNs is the ability to use a population count (popcount) for summation: an operation that consumes half the lookup tables (LUTs) of the otherwise-throughput-optimal balanced adder tree~\cite{BNN_CNN_FINN}.
		
		\begin{equation}
		    y = f{\left(\sum_{n=1}^{N}{w_n x_n}\right)}
			\label{eq:synapse_normal}
		\end{equation}
		
		\begin{figure}
		    \centering
		    \begin{tikzpicture}[thick, circuit logic US, thick, node distance=2.5mm, label distance=2mm, decoration={markings, mark=at position 0.5 with {\node [font=\footnotesize] {/};}}, scale=0.8, every node/.style={scale=0.8}]
	
	\newcommand{\updownkinklen}{2mm}
	\newcommand{\leftkinklen}{2.5mm}
	\newcommand{\upperinputlen}{2.5mm}
	\newcommand{\lowerinputlen}{2.5mm}
	\newcommand{\rightkinklen}{2.5mm}
	\newcommand{\outputlen}{7.5mm}
	
	\tikzstyle {xnor} = [xnor gate, point right, inputs={nn}]
	\tikzstyle {lut} = [rectangle, minimum width=12mm, minimum height=7.5mm, align=center, draw]
	\tikzstyle {sum} = [circle, minimum width=7.5mm, minimum height=7.5mm, align=center, draw]
	\tikzstyle {arrow} = [->, >=stealth]
	
	\node (lut3) [lut] {2-LUT};
	\node (lut2) [lut, above=of lut3, white] {2-LUT};
	\node (pruned) [xnor] at (lut2) {};
	\node (lut1) [lut, above=of lut2] {2-LUT};
	\node (lut4) [lut, below=of lut3, white] {};
	\node at ([yshift=1mm]lut4) {$\vdots$};
	\node (lut5) [lut, below=of lut4] {2-LUT};
	\node (sum2) [sum, right=of lut3, xshift=2.5mm] {$\Sigma$};
	
	\node (to) [left=of lut3, xshift=-24.5mm] {};
	\draw [fill=black] ([xshift=-5mm, yshift=-5mm]to.center) -- ([xshift=-5mm, yshift=5mm]to.center) -- ([yshift=5mm]to.center) -- ([yshift=7.5mm]to.center) -- ([xshift=5mm]to.center) -- ([yshift=-7.5mm]to.center) -- ([yshift=-5mm]to.center) -- cycle;
	
	\node (sum1) [sum, left=of to, xshift=-11.5mm] {$\Sigma$};
	\node (xnor3) [xnor, left=of sum1, xshift=-2.5mm] {};
	\node (xnor1) [xnor] at (lut1 -| xnor3) {};
	\node (xnor2) [xnor] at (lut2 -| xnor3) {};
	\node at ([yshift=1mm]lut4 -| xnor3) {$\vdots$};
	\node (xnor5) [xnor] at (lut5 -| xnor3) {};
	
	\draw [red] ([xshift=-5mm, yshift=-5mm]pruned.center) -- ([xshift=5mm, yshift=5mm]pruned.center);
	\draw [red] ([xshift=-5mm, yshift=5mm]pruned.center) -- ([xshift=5mm, yshift=-5mm]pruned.center);
	
	\draw (xnor1.input 1 -| lut1.west) -- ++(left:\leftkinklen) -- ++(up:\updownkinklen) -- ++(left:\upperinputlen) node [left, anchor=east] {$\tilde{x}_1^{\left(1\right)} = x_1$};
	\draw (xnor1.input 2 -| lut1.west) -- ++(left:\leftkinklen) -- ++(down:\updownkinklen) -- ++(left:\lowerinputlen) node [left, anchor=east] {$\tilde{x}_2^{\left(1\right)}$};
	\draw (xnor3.input 1 -| lut3.west) -- ++(left:\leftkinklen) -- ++(up:\updownkinklen) -- ++(left:\upperinputlen) node [left, anchor=east] {$\tilde{x}_1^{\left(2\right)} = x_3$};
	\draw (xnor3.input 2 -| lut3.west) -- ++(left:\leftkinklen) -- ++(down:\updownkinklen) -- ++(left:\lowerinputlen) node [left, anchor=east] {$\tilde{x}_2^{\left(2\right)}$};
	\draw (xnor5.input 1 -| lut5.west) -- ++(left:\leftkinklen) -- ++(up:\updownkinklen) -- ++(left:\upperinputlen) node [left, anchor=east] {$\tilde{x}_1^{\left(\tilde{N}\right)} = x_{N}$};
	\draw (xnor5.input 2 -| lut5.west) -- ++(left:\leftkinklen) -- ++(down:\updownkinklen) -- ++(left:\lowerinputlen) node [left, anchor=east] {$\tilde{x}_2^{\left(\tilde{N}\right)}$};
	\draw [arrow] (lut1.east) -- ++(right:\rightkinklen) -- (sum2);
	\draw [arrow] (lut3.east) -- ++(right:\rightkinklen) -- (sum2);
	\draw [arrow] (lut5.east) -- ++(right:\rightkinklen) -- (sum2);
	\draw [line width=1.5pt] (sum2.east) -- ++(right:\outputlen);
	\draw ([xshift=-1mm, yshift=-1mm]$(sum2.east) + (\outputlen/2,0)$) -- ([xshift=1mm, yshift=1mm]$(sum2.east) + (\outputlen/2,0)$) node [midway, above, anchor=west, rotate=90, xshift=1mm] {$\left\lceil\log_2{\left(\tilde{N}\right)}\right\rceil$};
	
	\draw (xnor1.input 1) -- ++(left:\leftkinklen) -- ++(up:\updownkinklen) -- ++(left:\upperinputlen) node [left, anchor=east] {$x_1$};
	\draw (xnor1.input 2) -- ++(left:\leftkinklen) -- ++(down:\updownkinklen) -- ++(left:\lowerinputlen) node [left, anchor=east] {$w_1$};
	\draw (xnor2.input 1) -- ++(left:\leftkinklen) -- ++(up:\updownkinklen) -- ++(left:\upperinputlen) node [left, anchor=east] {$x_2$};
	\draw (xnor2.input 2) -- ++(left:\leftkinklen) -- ++(down:\updownkinklen) -- ++(left:\lowerinputlen) node [left, anchor=east] {$w_2$};
	\draw (xnor3.input 1) -- ++(left:\leftkinklen) -- ++(up:\updownkinklen) -- ++(left:\upperinputlen) node [left, anchor=east] {$x_3$};
	\draw (xnor3.input 2) -- ++(left:\leftkinklen) -- ++(down:\updownkinklen) -- ++(left:\lowerinputlen) node [left, anchor=east] {$w_3$};
	\draw (xnor5.input 1) -- ++(left:\leftkinklen) -- ++(up:\updownkinklen) -- ++(left:\upperinputlen) node [left, anchor=east] {$x_{N}$};
	\draw (xnor5.input 2) -- ++(left:\leftkinklen) -- ++(down:\updownkinklen) -- ++(left:\lowerinputlen) node [left, anchor=east] {$w_{N}$};
	\draw [arrow] (xnor1.output) -- ++(right:\rightkinklen) -- (sum1);
	\draw [arrow] (xnor2.output) -- ++(right:\rightkinklen) -- (sum1);
	\draw [arrow] (xnor3.output) -- ++(right:\rightkinklen) -- (sum1);
	\draw [arrow] (xnor5.output) -- ++(right:\rightkinklen) -- (sum1);
	\draw [line width=1.5pt] (sum1.east) -- ++(right:\outputlen);
	\draw ([xshift=-1mm, yshift=-1mm]$(sum1.east) + (\outputlen/2,0)$) -- ([xshift=1mm, yshift=1mm]$(sum1.east) + (\outputlen/2,0)$) node [midway, above, anchor=west, rotate=90, xshift=1mm] {$\left\lceil\log_2{\left(N\right)}\right\rceil$};
	
\end{tikzpicture}
		    \caption{
		        BNN to LUTNet architectural transformation for a single channel, mirroring the replacement of \eqref{eq:synapse_normal} with \eqref{eq:synapse_lutnet}.
		        Activation function blocks are not shown, but follow the adders.
		        $\tilde{N}$ lookup tables (here, 2-LUTs) substitute $N$ XNOR gates. $\tilde{N} \ll N$ is achieved via pre-substitution pruning, represented by the removal---\emph{i.e.} lack of LUT substitution---of the second XNOR gate.
		        LUT inputs $\tilde{x}_1^{\left(n\right)}~\forall n$ are connected to preserve the pruned BNN's structure.
		        In this case, LUTNet's weights are encoded in its LUT masks, thus they do not appear as inputs.
            }
		    \label{fig:transform}
		\end{figure}
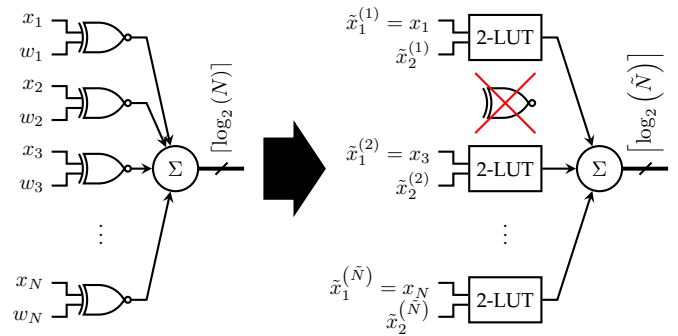
		
		No matter how simple these multiplications become, however, all of the products still need to be summed.
		In modern networks, $N$ commonly reaches numbers in the thousands~\cite{ALEXNET}.
		To tackle this, we propose the replacement of \eqref{eq:synapse_normal} with the specifically FPGA-inspired function \eqref{eq:synapse_lutnet}, wherein the activation function is unchanged but each product is replaced with an \emph{arbitrary} term-specific Boolean function $g_n : \left\{-1,1\right\}^K \to \left\{-1,1\right\}$.
		The input to this function is a vector $\tilde{\boldsymbol{x}}^{\left(n\right)}$ whose elements are any $K$ components of the original input vector $\boldsymbol{x}$, \emph{i.e.} $\tilde{\boldsymbol{x}}^{\left(n\right)} = \boldsymbol{S}_n\boldsymbol{x}$ for some binary selection matrix $\boldsymbol{S}_n \in \left\{0,1\right\}^{K \times N}$ with ${\left\lVert\boldsymbol{S}_n\right\rVert}_\infty = 1$.
		Since its inputs and outputs are binary, each $g_n$ maps directly to a single $K$-LUT.
		BNNs are a special case of this function: they are recoverable when $K = 1$ and $\tilde{N} = N$, with $\boldsymbol{S}_n$ being the row vector with the $n$\textsuperscript{th} element equal to one and all others zero.
		An example of the resultant architectural transformation---excluding blocks for $f$, which are common to both approaches---is given in Fig.~\ref{fig:transform}.
		
		\begin{equation}
			y = f{\left(\sum_{n=1}^{\tilde{N}} g_n{\left(\tilde{\boldsymbol{x}}^{\left(n\right)}\right)}\right)}
			\label{eq:synapse_lutnet}
		\end{equation}
		
		
		Notice that, while in \eqref{eq:synapse_normal} each element of $\boldsymbol{x}$ only participates in a single summation term, in \eqref{eq:synapse_lutnet} each can participate in many terms.
		The intuition here is that inputs can be arranged such that $\tilde{N} \ll N$ for comparable accuracy via network pruning, dramatically reducing the sizes of the required popcount trees.
		Our experiments demonstrate that this is indeed the case.
		
\begin{figure}
		    \centering
		    \begin{tikzpicture}[thick, circuit logic US, thick, node distance=2.5mm, label distance=2mm, decoration={markings, mark=at position 0.5 with {\node [font=\footnotesize] {/};}}, scale=0.7, every node/.style={scale=0.8}]
	
	\newcommand{\updownkinklen}{2.3mm}
	\newcommand{\leftkinklen}{2.5mm}
	\newcommand{\upperinputlen}{2.5mm}
	\newcommand{\lowerinputlen}{2.5mm}
	\newcommand{\rightkinklen}{2.5mm}
	\newcommand{\outputlen}{7.5mm}
	
	\tikzstyle {xnor} = [xnor gate, point right, inputs={nn}]
	\tikzstyle {lut} = [rectangle, minimum width=12mm, minimum height=7.5mm, align=center, draw]
	\tikzstyle {sum} = [circle, minimum width=7.5mm, minimum height=7.5mm, align=center, draw]
	\tikzstyle {arrow} = [->, >=stealth]
	
	\node (lut3) [lut] {2-LUT};
	\node (lut2) [lut, above=of lut3, yshift=0mm, white] {};
	\node (lut1) [lut, above=of lut3, yshift=5mm] {2-LUT};
	\node (lut4) [below=of lut3, yshift=0mm]{$\vdots$};
	\node (lut5) [lut, below=of lut4, yshift=-2.5mm] {2-LUT};
	\node (sum1) [sum, right=of lut3, xshift=2.5mm] {$\Sigma$};
	\node (sum2) [sum, right=of sum1, xshift=3mm] {$\Sigma$};
	\node (reg1) [right=of sum2, xshift=3mm] {};
	
	
	\draw ([yshift=1mm]lut1.west) -- ++(left:\leftkinklen) -- ++(up:\updownkinklen) -- ++(left:\upperinputlen) node [left, anchor=east] {$\tilde{x}^{\left(1,1\right)}_{1}, \tilde{x}^{\left(1,2\right)}_{1}, \cdots$};
	\draw ([yshift=1mm]lut3.west) -- ++(left:\leftkinklen) -- ++(up:\updownkinklen) -- ++(left:\upperinputlen) node [left, anchor=east] {$\tilde{x}^{\left(2,1\right)}_{1}, \tilde{x}^{\left(2,2\right)}_{1}, \cdots$};
	\draw ([yshift=1mm]lut5.west) -- ++(left:\leftkinklen) -- ++(up:\updownkinklen) -- ++(left:\upperinputlen) node [left, anchor=east] {$\tilde{x}^{\left(\nicefrac{\tilde{N}}{T},1\right)}_{1}, \tilde{x}^{\left(\nicefrac{\tilde{N}}{T},2\right)}_{1}, \cdots$};
	
	
	\draw [arrow] (lut1.east) -- ++(right:\rightkinklen) -- (sum1);
	\draw [arrow] (lut3.east) -- ++(right:\rightkinklen) -- (sum1);
	\draw [arrow] (lut5.east) -- ++(right:\rightkinklen) -- (sum1);
	\draw [arrow, line width=1.5pt] (sum1.east) -- (sum2.west);
	\draw ([xshift=-2mm, yshift=-1mm]$(sum1.east) + (\outputlen/2,0)$) -- ([yshift=1mm]$(sum1.east) + (\outputlen/2,0)$) node [midway, above, anchor=west, rotate=90, xshift=1mm] {$\left\lceil\log_2{\left(\nicefrac{\tilde{N}}{T}\right)}\right\rceil$};
	
	

    
	\node (bram) [above=of lut1, xshift=-50mm, yshift=5mm] {RAM};
	\draw ([xshift=-10mm, yshift=-5mm]bram.center) -- ([xshift=-10mm, yshift=5mm]bram.center) -- ([xshift=10mm, yshift=5mm]bram.center) -- ([xshift=10mm, yshift=-5mm]bram.center) -- cycle;
	
	
	\node (cross1) at (bram.center) [yshift=-4.5mm] {};
	\node [circle, fill=black, inner sep=1.5pt] (cross2) at (cross1 |- lut1.south west) [yshift=1mm] {};
	\node [circle, fill=black, inner sep=1.5pt] (cross3) at (cross1 |- lut3.south west) [yshift=1mm] {};
	\node (cross4) [below=of cross3.center, yshift=-1mm] {};
	\node (cross5) [below=of cross4.center, yshift=-4mm] {};
	\node at (cross1 |- lut4.center){$\vdots$};
	\node (cross6) at (cross1 |- lut5.south west) [yshift=1mm] {};
	
	\draw [line width=1.5pt] (cross1.center) -- (cross2.center) node (mark_location) [midway, anchor=center] {};
	\draw ([xshift=-1mm, yshift=-1mm]$(mark_location.center)$) -- ([xshift=1mm, yshift=1mm]$(mark_location.center)$) node [midway, above, anchor=east, rotate=0, xshift=0mm] {$\nicefrac{\tilde{N}}{T}$};
	\draw [line width=1.5pt] (cross2.center) -- (cross3.center);
	\draw [line width=1.5pt] (cross3.center) -- (cross4.center);
	\draw (cross5.center) -- (cross6.center);
	\draw (cross2.center) -- ([xshift=-2.5mm,yshift=1mm]lut1.south west) node [midway, below, anchor=north] {$\tilde{p}_1^{\left(1,1\right)}, \tilde{p}_1^{\left(1,2\right)}, \cdots$} |- ([yshift=-1mm]lut1.west);
	\draw (cross3.center) -- ([xshift=-2.5mm,yshift=1mm]lut3.south west) node [midway, below, anchor=north] {$\tilde{p}_1^{\left(2,1\right)}, \tilde{p}_1^{\left(2,2\right)}, \cdots$} |- ([yshift=-1mm]lut3.west);
	\draw (cross6.center) -- ([xshift=-2.5mm,yshift=1mm]lut5.south west) node [midway, below, anchor=north] {$\tilde{p}_1^{\left(\nicefrac{\tilde{N}}{T},1\right)}, \tilde{p}_1^{\left(\nicefrac{\tilde{N}}{T},2\right)}, \cdots$} |- ([yshift=-1mm]lut5.west);
	
	
	\node [circle, fill=black, inner sep=1.5pt] (cross7) [right=of reg1, xshift=-0.5mm] {};
	\node (cross8) [above=of cross7, yshift=5mm] {};
	\draw ([xshift=-3mm, yshift=-4mm]reg1.center) -- ([xshift=-3mm, yshift=3mm]reg1.center) -- ([xshift=3mm, yshift=3mm]reg1.center) -- ([xshift=3mm, yshift=-4mm]reg1.center) -- cycle;
	\draw ([xshift=-1.5mm, yshift=-4mm]reg1.center) -- ([xshift=0mm, yshift=-2.5mm]reg1.center) -- ([xshift=1.5mm, yshift=-4mm]reg1.center);
	\draw [arrow, line width=1.5pt] (sum2.east) -- ([xshift=-3mm]reg1.center);
	\draw [line width=1.5pt] ([xshift=3mm]reg1.center) -- ++(right:5mm);
	\draw [line width=1.5pt] (cross7.center) -- (cross8.center);
	\draw [arrow, line width=1.5pt] (cross8.center) -| (sum2.north);
	
\end{tikzpicture}
		    \caption{
		        Tiled version of the LUTNet architecture shown in Fig.~\ref{fig:transform}.
		        $\nicefrac{\tilde{N}}{T}$ 2-LUTs substitute Fig.~\ref{fig:transform}'s $\tilde{N}$ 2-LUTs, with one of each of the former's inputs now connected to on-chip RAM.
		        LUT masks and parameters stored in memory are both learned during training.
            }
            \label{fig:transform_tm}
        \end{figure}
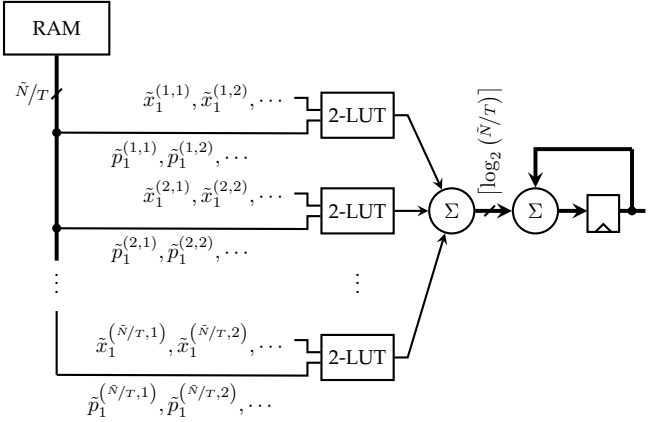

		The hardware realization of \eqref{eq:synapse_lutnet} requires one-to-one $g_n \to \text{LUT}$ binding.
		Given the sizes of today's DNN models, this either limits LUTNet's deployment to only a handful of layers or, in scenarios where throughput and energy efficiency are of paramount importance, \emph{e.g.} for cloud-based computing~\cite{ASAP}, makes whole-network implementation expensive.
		By sacrificing a subset of LUT inputs and feeding them with supplementary parameters from random-access memory (RAM), we can trade off throughput and efficiency for additional accuracy.
		To achieve this, we transform \eqref{eq:synapse_lutnet} into \eqref{eq:synapse_tm_lutnet}, partitioning channel inputs over $T$ non-overlapping \emph{tiles}~\cite{Parhi}.
		The implementation of \eqref{eq:synapse_tm_lutnet} is exemplified in Fig.~\ref{fig:transform_tm}.
		
		\begin{equation}
			y = f{\left(\sum_{t=1}^{T}\sum_{m=1}^{\nicefrac{\tilde{N}}{T}} g_m{\left(\tilde{\boldsymbol{x}}^{\left(m,t\right)},\tilde{\boldsymbol{p}}^{\left(m,t\right)}\right)}\right)}
			\label{eq:synapse_tm_lutnet}
		\end{equation}
		
		In moving from \eqref{eq:synapse_lutnet} to \eqref{eq:synapse_tm_lutnet}, bivariate functions $g_m : \left\{-1,1\right\}^{K - P} \times \left\{-1,1\right\}^P \to \left\{-1,1\right\}$ replace every $g_n$, each of which is used $T$ times per calculation of $y$.
		Now, $K - P$ inputs for some $P < K$ are available per node for connection to $\boldsymbol{x}$; the remaining $P$ are used to receive additional learned parameters, $\tilde{\boldsymbol{p}}$, streamed in from RAM.
		These parameters effectively allow runtime selection between $2^P$ candidate $K - P : 1$ Boolean operations per $g_m$.
		Note that \eqref{eq:synapse_tm_lutnet} is a strict generalization of \eqref{eq:synapse_lutnet}: when $T = 1$ and $P = 0$, the former reverts to the latter.
		Comparing Fig.~\ref{fig:transform_tm} with the equivalent unrolled implementation shown in Fig.~\ref{fig:transform}, $K$-LUT requirements have been reduced by a factor of $T$ and the popcount tree has been thinned, with small overheads introduced due to the need for RAM and an accumulator.
		
		Our aim in proposing these new inference node functions is to play to the strengths of FPGA soft logic.
		While a LUT is capable of performing an arbitrary \emph{nonlinear Boolean} function, traditional DNNs are based around
		\begin{enumerate*}[label=(\roman*)]
		    \item \emph{near-linear}\label{item:near-lin}
		    \item \emph{high-precision}\label{item:high-prec}
		\end{enumerate*}
		functions: almost the exact opposite of the architecture's forte.
		Innovations such as BNNs have addressed \ref{item:near-lin}, by reducing precision~\cite{CSUR, STYLIANOS_TOOLFLOWS, SURV_FPGA_BASED_NEURAL_NETWORK_ACC}; we address both \ref{item:near-lin} and \ref{item:high-prec} by also embracing the nonlinearity of the LUT.
		Herein, we make the following novel contributions.
		\begin{itemize}
			\item
			    We introduce LUTNet, the first neural network architecture featuring $K$-LUTs as inference operators.
			    Since each $K$-LUT is capable of performing an arbitrary Boolean operation on up to $K$ inputs, LUTNet's logic density is much greater than that of BNNs.
			\item
			    We propose a training regime resulting in the conversion of a BNN architecture from a dense array of XNOR gates into a sparse network of arbitrary $K$-input functions directly mappable onto $K$-LUTs.
			\item
			    We extend our training program to natively support network tiling, allowing inference nodes to be shared between operations both within and across channels.
			    This facilitates whole-network LUTNet deployment on current-generation FPGAs.
			\item
			    We empirically demonstrate the effects of LUTNet's increased logic density on area efficiency and accuracy.
			    We also experimentally explore the associated energy and training efficiency impacts.
			    Our results for unrolled, 4-LUT-based inference operators reveal area compression of 2.08$\times$ and 1.90$\times$ for the CNV network~\cite{BNN_CNN_FINN} classifying the CIFAR-10 dataset~\cite{CIFAR10} and AlexNet~\cite{ALEXNET} classifying ImageNet~\cite{IMAGENET}, respectively, against an unrolled and losslessly pruned implementation of ReBNet~\cite{BNN_CNN_REBNET_FCCM}, the state-of-the-art BNN, while achieving comparable accuracy.
			\item
			    We comprehensively explore the $\left(P,T\right)$ parameter space offered by our tiling-friendly architecture, finding that, while area and energy efficiency gains over ReBNet are less dramatic than in the unrolled case, we still achieve improvements in both metrics: up to 1.28$\times$ and 1.57$\times$, respectively.
			\item
			    We provide an open-source release\footnotemark of LUTNet for the community to use and build upon.
		\end{itemize}
		
		\footnotetext{\texttt{https://github.com/awai54st/LUTNet}}
		
		A preliminary version of this work appeared in the proceedings of the 27th IEEE International Symposium on Field-Programmable Custom Computing Machines (FCCM)~\cite{LUTNET}.
		The early-stage implementation we described in that paper required the complete unrolling of network layers in order to realize the LUTNet architecture, limiting scalability.
		In this article, we describe how the sacrifice of $K$-LUT inputs can be used to enable tiling.
		This adds additional parameters---the number of weight inputs per LUT and tiling factors---to our design space which we empirically explore, finding favorable combinations while enabling whole-network deployment on today's FPGAs.
		Finally, we provide a documented, open-source release\footnotemark[\value{footnote}] of LUTNet's training and implementation code.
		
	\section{Related Work}
	
	    \subsection{Quantization}
	    
    	    The authors of early BNN publications, such as BinaryConnect~\cite{BNN_CNN_BinaryConnect} and BinaryNet~\cite{BNN_CNN_BinaryNet}, proposed network training with binary weights and activations (channel inputs and outputs) used for forward propagation.
    	    High-precision formats---most commonly IEEE-754 single-precision floating point, used to approximate reals $\mathbb{R}$---are always used for backward propagation; this is essential in order for stochastic gradient descent to work well~\cite{BNN_CNN_BinaryConnect,AARON_ZHAO_FPT,AARON_ZHAO_NIPS}.
            Tang \emph{et al.} showed that training from scratch with binarized forward propagation is significantly slower than through the consistent use of high-precision data, however; learning rates some 100$\times$ lower are required than in the all-real case~\cite{BNN_CNN_BINARY_CONSTRIANED_TRAINING}.
            Furthermore, binary forward propagation results in the majority of real-valued weights being close to either $-1$ or $1$, while a spread across $\left[-1, 1\right]$ is required to facilitate fine-grained pruning~\cite{PRU_CNN_TRAIN_PRUNE_RETRAIN}.
            
            Improving upon BinaryNet's data representation, Rastegari \emph{et al.}'s BWN features layer-wise trainable scaling factors $\boldsymbol{\alpha}$ used in order to increase BNN expressiveness~\cite{BNN_CNN_XNOR-Net}.
            During training, each $\alpha_l \in \mathbb{R}$ assumes the mean value of layer $l$'s weights.
            When inferencing, this is multiplied with the layer's popcount results, compensating for some of the information lost to binarization and increasing accuracy.
    
            Tang \emph{et al.}~\cite{BNN_CNN_BINARY_CONSTRIANED_TRAINING} and the authors of ABC-Net~\cite{BNN_CNN_ABC-Net} and ReBNet demonstrated the alleviation of information loss from binarization through the approximation of real-valued weights as linear combinations of multiple binary values.
            This is achieved via \emph{residual binarization}, a scheme in which each bit is the binarized residual error of its predecessor.
            Each bit $b$ is associated with a trainable scaling factor $\gamma_b \in \mathbb{R}$, representing its relative importance.
            When quantizing, each weight $\hat{w} \in \mathbb{R}$ is approximated as $B$ binary weights $w_b = \text{sign}{\left(\epsilon_b\right)}$, as shown in \eqref{eqn:residual_binarisation}, wherein $\epsilon_b$ is the $b$\textsuperscript{th} bit's residual error.
            During training, each $\gamma_b$ is updated to minimize the total error.
            While accuracy was found to be positively correlated with $B$, diminishing returns were seen; little improvement was observed for $B > 2$.
            
            \begin{equation}
                \begin{gathered}
                    \hat{w} = \sum^B_{b=1}{\gamma_b~w_b} \\
                    \epsilon_b = \epsilon_{b-1}-\gamma_{b-1}~\text{sign}{\left(\epsilon_{b-1}\right)}
                \end{gathered}
                \label{eqn:residual_binarisation}
            \end{equation}
        
        \subsection{Pruning}
	    
        	Use of fine-grained pruning effectively adds zero to the set of possible binary weight values, resulting in a ternary representation.
        	Ternarization has been shown by the authors of many works to deliver significantly higher accuracy than binarization~\cite{TNN_CNN_TWN,TNN_CNN_TTQ,TNN_CNN_BENGIO}.
            Pruning also promotes regularization, reducing overfitting~\cite{PRU_CNN_STRUCTURED_SPARSITY}.
            The latter is particularly relevant to this work since the use of $K$-LUTs as inference operators greatly increases potential network complexity.
            
            In order to promote pruning, Han \emph{et al.} proposed training with the $l_2$ sparsification regularizer in \eqref{eqn:cnn_sparsity_regulariser}~\cite{PRU_CNN_TRAIN_PRUNE_RETRAIN}.
            During backward propagation, $\Omega$ influences training loss, inducing weights carrying low significance to descend towards zero.
            $\lambda$, $L$, and $C$ are the regularization factor, number of layers, and number of channels per layer, respectively.
            $\hat{\boldsymbol{w}}^{\left(l,c\right)}$ denotes the real-valued weight vector of layer $l$'s channel $c$.
            
            \begin{equation}
                \Omega = \lambda\sqrt{\sum^L_{l=1}{\sum^C_{c=1}{{\left(\hat{\boldsymbol{w}}^{\left(l,c\right)}\right)}^2}}}
                \label{eqn:cnn_sparsity_regulariser}
            \end{equation}
        
        \subsection{Tiling}
	    
            Rather than computing the outputs of entire network layers in parallel, many authors have explored the partitioning of weight matrices and corresponding input activations into non-overlapping tiles.
            Computing tiles sequentially typically reduces throughput and increases latency but increases opportunities for resource sharing, facilitating area and power consumption reductions.
            Zhang \emph{et al.} proposed tiling along both input and output channels~\cite{NR_CNN_FXP_OPTIM_LOOP_JASON_CONG}.
            While authors including Ma \emph{et al.} have also proposed tiling across other dimensions, such as within convolutional windows, these have been shown to typically provide fewer opportunities for resource sharing than intra-channel strategies~\cite{NR_CNN_FXP_OPTIM_LOOP}.
    
        All of the aforementioned proposals are complementary to our approach, which uses $K$-LUTs in their full generality.
		We integrate this prior work through the use of high-precision training, fine-grained pruning, layer-wise scaling factors, and residual binarization, combining it with the key LUTNet novelty to achieve state-of-the-art performance significantly more cheaply than previously reported in the literature.
		We also support tiling over input and output channels in order to enable whole-network deployment.
		
		\subsection{Architectural Modification}
	    
    		Authors including Boutrous \emph{et al.}~\cite{BETZ_FPGA19} and Rasoulinezhad \emph{et al.}~\cite{PIRDSP_FCCM19} have proposed modifications to FPGA fabrics to suit the implementation of low-precision dot product operators, including additional carry chains and finer-grained multiplier and adder fracturability.
    		We take the opposite approach: rather than changing the hardware to suit existing DNN arithmetic, we change the arithmetic to suit the FPGA platforms currently on the market.
    		
    		Given that there is evidence showing that neural networks perform classification by simply memorizing their training data, many find it surprising that they can generalize on unseen test data~\cite{ZHANG_RETHINKING_GENERALISATION}.
            To explain this phenomenon, Chatterjee proposed a deep network architecture entirely constituting small memory blocks performing table lookups, showing that DNN generalization can be achieved by memorization alone~\cite{CHATTERJEE_MEMORIZATION}.
            In contrast, we approach LUT-based DNN inference from a hardware-oriented motivation: given existing FPGA LUTs, we seek to achieve the best area-accuracy tradeoffs possible.
            Unlike Chatterjee's software prototype, we present details of hardware implementations that beat state-of-the-art BNN inference designs.
            Complementarily to our LUT-based architecture, we also integrate commonly used DNN components including convolution and pooling to improve performance.
            
	\section{Network Construction and Training}
	\label{sec:training}
	    
	    LUTNet's initialization comprises three successive stages: training, pruning, and \emph{``logic expansion"} (XNOR to $K$-LUT conversion), with each of the latter two including a retraining phase.
	    All three phases were implemented with TensorFlow.
	    While our training and pruning stages are fairly standard, the final phase---logic expansion---encompasses the key novelty of our approach.
	    
	    \subsection{Training}
	    \label{sec:training_training}
			
			In order to both expedite learning and facilitate later pruning, our first step is to train the chosen network model using high-precision data during both forward and backward propagation.
            Layer-wise scaling factors $\boldsymbol{\alpha}$ are learned during this stage along with weights, and sparsification is induced through the use of the $l_2$ regularizer in \eqref{eqn:cnn_sparsity_regulariser} with $\lambda = \num{5e-7}$ as suggested by Tang \emph{et al.}~\cite{BNN_CNN_BINARY_CONSTRIANED_TRAINING}.
	    
	    \subsection{Pruning}
	    \label{sec:training_pruning}
	    
	        Following high-precision training, fine-grained pruning is conducted through the application of threshold $\theta$ on each weight $\hat{w}$, as shown in \eqref{eqn:threshold}.
	        $\theta$ exposes a continuum between area occupancy and accuracy: the higher its value, the more weights are pruned away.
	        
            \begin{equation}
                \begin{gathered}
                    \hat{w} \gets \begin{cases}
                        \hat{w} & \text{if}~\lvert \hat{w} \rvert > \theta \\
                        0       & \text{otherwise}
                    \end{cases}
                \end{gathered}
                \label{eqn:threshold}
            \end{equation}
	        
	        Once pruned, the network is binarized following the scheme shown in \eqref{eqn:residual_binarisation}, after which it is retrained in order to recover some of the induced accuracy loss.
	        Due to the diminishing returns previously found when applying residual binarization~\cite{BNN_CNN_BINARY_CONSTRIANED_TRAINING,BNN_CNN_ABC-Net,BNN_CNN_REBNET_FCCM}, we used $B = 2$ (two-level binarization) consistently.

	    \subsection{Logic Expansion}
	    \label{sec:training_expansion}
	        
			At this point, we have obtained a residual-binarized ternary neural network with non-zero-weighted operators implemented as XNORs.
	        It is from here that we depart from the standard BNN approach.
	        
	        Each XNOR gate is replaced with a $K$-LUT, whose first input $\tilde{x}^{\left(m,t\right)}_1$ is assigned to preserve the original connection, thereby retaining the pruned BNN's structure.
	        If $K - P > 1$, the $K - P - 1$ subsequent inputs to each LUT are then randomly selected from channel inputs within the same convolutional window as $\tilde{x}^{\left(m,t\right)}_1$, ensuring that the window shape remains unchanged.
	        We additionally constrain their selection such that each channel input is connected at most once to each LUT.
		    Where $P > 0$, each LUT's final $P$ inputs are connected to a $P$-bit-wide memory element, $\tilde{\boldsymbol{p}}^{\left(m,t\right)}$.
		    
		    Given the aforementioned restrictions on the sources of additional LUT connections, it is possible that, with large $K$ and low $P$, there will be insufficient (\emph{i.e.} $< K - P - 1$) candidate signals with which to saturate the LUTs.
		    If this does happen, $K - P$ should be reduced in order not to waste inputs.
		    In practise, this scenario is unlikely to manifest: DNNs are complex and sensible choices of $K$ are related to the size of the physical LUTs on the target device, which is typically small.
		    We did not encounter this issue for any of the networks we experimented with.
	        
	        The form of the inference function proposed in \eqref{eq:synapse_tm_lutnet} is defined on the binary domain $\left\{-1,1\right\}^{\tilde{N}}$.
            In common with quantization-inspired networks such as BNNs, this causes difficulty for training algorithms designed to operate on real vectors ${\mathbb R}^{\tilde{N}}$, specifically in the backward propagation of derivatives. 
            Our approach to this problem is to define an \emph{interpolating extension} of the function $g_m : \left\{-1,1\right\}^{K-P} \times \left\{-1,1\right\}^{P} \to \left\{-1,1\right\}$, \emph{i.e.} a function $\hat{g}_m : \mathbb{R}^{K-P} \times \mathbb{R}^{P} \to \mathbb{R}$ such that $\hat{g}_m{\left(\tilde{\boldsymbol{x}}^{\left(m,t\right)},\tilde{\boldsymbol{p}}^{\left(m,t\right)}\right)} = g_m{\left(\tilde{\boldsymbol{x}}^{\left(m,t\right)},\tilde{\boldsymbol{p}}^{\left(m,t\right)}\right)}$ for every $\tilde{\boldsymbol{x}}^{\left(m,t\right)}$ and $\tilde{\boldsymbol{p}}^{\left(m,t\right)}$ in the domain of $g_m$.
            There are many such functions.
            Of them, we prefer those that are as smooth as possible, allowing training optimization methods to perform well, and that form a good interpolation in the sense that, if $g_m$ remains constant when a Boolean input flips, so does $\hat{g}_m$.
            A natural choice for the extension is a Lagrange interpolating polynomial, leading to the form we use in \eqref{eqn:Lagrange_interpl}.
            
            \begin{equation}
                \hat{g}_m{\left(\hat{\tilde{\boldsymbol{x}}}^{\left(m,t\right)},\hat{\tilde{\boldsymbol{p}}}^{\left(m,t\right)}\right)} = \sum_{\boldsymbol{d} \in \left\{-1,1\right\}^K}{\left(\hat{c}_{\boldsymbol{d}} \prod_{k = 1}^K{\left(\begin{bmatrix}\hat{\tilde{\boldsymbol{x}}}^{\left(m,t\right)} \\ \hat{\tilde{\boldsymbol{p}}}^{\left(m,t\right)}\end{bmatrix}_k - d_k\right)}\right)}
                \label{eqn:Lagrange_interpl}
            \end{equation}
            
            \noindent This expands as shown in \eqref{eqn:Lagrange_interpl_expand} for $K > 0$ and $P \geq 0$, with each polynomial comprising $2^K$ trainable parameters $\hat{\boldsymbol{c}}$.
            
            \begin{multline}
                \hat{g}_m{\left(\hat{\tilde{\boldsymbol{x}}}^{\left(m,t\right)},\hat{\tilde{\boldsymbol{p}}}^{\left(m,t\right)}\right)} =  \\
                \begin{cases}
                    \begin{aligned}
                        &\hat{c}_{\left(-1\right)}{\left(\hat{\tilde{x}}^{\left(m,t\right)}_1 + 1\right)}  \\
                        &+ \hat{c}_{\left(1\right)}{\left(\hat{\tilde{x}}^{\left(m,t\right)}_1 - 1\right)}
                    \end{aligned}	& \text{if}~K = 1,~P = 0    \\
                    \\
                    \begin{aligned}
                        &\hat{c}_{\left(-1,-1\right)}{\left(\hat{\tilde{x}}^{\left(m,t\right)}_1 + 1\right)\left(\hat{\tilde{x}}^{\left(m,t\right)}_2 + 1\right)}	\\
                        &+ \hat{c}_{\left(-1,1\right)}{\left(\hat{\tilde{x}}^{\left(m,t\right)}_1 + 1\right)\left(\hat{\tilde{x}}^{\left(m,t\right)}_2 - 1\right)}	\\
                        &+ \hat{c}_{\left(1,-1\right)}{\left(\hat{\tilde{x}}^{\left(m,t\right)}_1 - 1\right)\left(\hat{\tilde{x}}^{\left(m,t\right)}_2 + 1\right)}	\\
                        &+ \hat{c}_{\left(1,1\right)}{\left(\hat{\tilde{x}}^{\left(m,t\right)}_1 - 1\right)\left(\hat{\tilde{x}}^{\left(m,t\right)}_2 - 1\right)}
                    \end{aligned}	& \text{if}~K = 2,~P = 0    \\
                    \\
                    \begin{aligned}
                        &\hat{c}_{\left(-1,-1\right)}{\left(\hat{\tilde{x}}^{\left(m,t\right)}_1 + 1\right)\left(\hat{\tilde{p}}^{\left(m,t\right)}_1 + 1\right)}	\\
                        &+ \hat{c}_{\left(-1,1\right)}{\left(\hat{\tilde{x}}^{\left(m,t\right)}_1 + 1\right)\left(\hat{\tilde{p}}^{\left(m,t\right)}_1 - 1\right)}	\\
                        &+ \hat{c}_{\left(1,-1\right)}{\left(\hat{\tilde{x}}^{\left(m,t\right)}_1 - 1\right)\left(\hat{\tilde{p}}^{\left(m,t\right)}_1 + 1\right)}	\\
                        &+ \hat{c}_{\left(1,1\right)}{\left(\hat{\tilde{x}}^{\left(m,t\right)}_1 - 1\right)\left(\hat{\tilde{p}}^{\left(m,t\right)}_1 - 1\right)}
                    \end{aligned}	& \text{if}~K = 2,~P = 1    \\
                    \\
                    \cdots  & \cdots
                \end{cases}
                \label{eqn:Lagrange_interpl_expand}
			\end{multline}
            
            Since connections are effectively remade from an unpruned BNN (Section~\ref{sec:training_training}), it makes sense to use those channel inputs' original weights, $\hat{\tilde{\boldsymbol{w}}}$, as a starting point for retraining.
	        The goal of our initialization process is to achieve the identity \eqref{eqn:retrain_function} holding true for all values of $\hat{\tilde{\boldsymbol{x}}}^{\left(m,t\right)}$, wherein $\left\{2,\cdots,K-P\right\}$ represents the set of reconnected channel inputs that were removed via pruning (Section~\ref{sec:training_pruning}), if any.
            
            \begin{equation}
                \hat{g}_m{\left(\hat{\tilde{\boldsymbol{x}}}^{\left(m,t\right)},\hat{\tilde{\boldsymbol{p}}}^{\left(m,t\right)}\right)} = \sum_{i \in \left\{1,\cdots,K-P\right\}}{\hat{\tilde{x}}_i^{\left(m,t\right)}\hat{w}_i^{\left(m,t\right)}}
                \label{eqn:retrain_function}
            \end{equation}
			
			\eqref{eqn:retrain_function} can be solved by matching the monomial coefficients with respect to $\hat{\tilde{\boldsymbol{x}}}^{\left(m,t\right)}$ between the left- and right-hand sides of the equation.
	        When $P = 0$, there are $2^K$ equations with $2^K$ unknowns, thus there is exactly one solution.
	        If $P > 0$, however, we have $2^{K - P}$ equations and $P + 2^K$ unknowns, and so there are infinitely many solutions.
	        For the latter scenario, we choose the most intuitive solution by initializing $\hat{\tilde{\boldsymbol{p}}}^{\left(m,t\right)}$ with values from $\hat{\tilde{\boldsymbol{w}}}$.
	        When $P \leq \nicefrac{K}{2}$, \emph{i.e.} the number of memory connections does not exceed the number of input channel weights to initialize from, we simply let $\hat{\tilde{p}}_i^{\left(m, t\right)} \leftarrow \hat{\tilde{w}}_i^{\left(m, t\right)}~\forall i \in \left\{1, \cdots, \nicefrac{K}{2}\right\}$.
	        In the case where $P > \nicefrac{K}{2}$, we treat the first $\nicefrac{K}{2}$ elements of $\hat{\tilde{p}}_i^{\left(m, t\right)}$ in the same way, while the remainder are selected at random: $\hat{\tilde{p}}_i^{\left(m,t\right)} \sim \left\{-1, 1\right\}~\forall i \in \left\{\nicefrac{K}{2} + 1, \cdots, P\right\}$.
	        
	        Once all $\hat{g}_m$ are initialized, our second and final retraining phase is conducted, whereafter the binarized training parameters $\boldsymbol{c} = \text{sign}{\left(\hat{\boldsymbol{c}}\right)}$ and $\tilde{\boldsymbol{p}}^{\left(m,t\right)} = \text{sign}{\left(\hat{\tilde{\boldsymbol{p}}}^{\left(m,t\right)}\right)}$ can be directly interpreted as the configuration mask of each $K$-LUT and contents of each memory element, respectively.
	        It should be noted that, in the case where previously pruned connections remade in the solution of \eqref{eqn:retrain_function} are again found to be of little importance, this phase will drive those connections' respective $\hat{\boldsymbol{c}}$ parameters towards zero.
	        As a result, the hardware synthesis that follows will reprune them.
            
            We elected to follow the procedure detailed above rather than training with $K$-LUTs from scratch due to the exponential relationship between $K$ and the number of trainable parameters $\hat{\boldsymbol{c}}$.
            Our early experiments revealed that training these from the outset, particularly prior to network pruning, causes both slow convergence and frequent overfitting due to the large numbers of local minima in the search space.
    	    High-precision training followed by pruning not only ensures fast convergence, it also brings the starting point of $K$-LUT learning closer to global minima, reducing the likelihood of overfitting.
                
	\section{Network Implementation}
        
\begin{figure}
			\centering
			\begin{tikzpicture}[scale=1.0, every node/.style={scale=0.77}]
    
    \begin{axis}[
		width=0.48\columnwidth,
		height=0.48\columnwidth,
		axis lines = middle,
        ymin=0,
        ymax=5.5,
        xmin=0,
        xmax=6.5,
        xtick={0, 1, ..., 6},
        ytick={1, 2, ..., 5},
        yticklabels={0, 1, ..., 4},
		xlabel={$K$},
		ylabel={$P$},
        x label style={at={(axis description cs:1.05,-0.02)},anchor=north},
        y label style={at={(axis description cs:-0.0,1.05)},anchor=east},
        ]
        \addplot [thick, only marks, mark=+, mark options={scale=1.0, color=red}] coordinates {(1,2) (2,2) (3,2) (4,2) (5,2) (6,2) (3,3) (4,3) (5,3) (6,3) (5,4) (6,4) (6,5)}; \label{plt:combo_search_k_w_kwlutnet}
        \addplot [thick, only marks, mark=x, mark options={scale=1.0, color=blue}] coordinates {(1,1) (2,1) (3,1) (4,1) (5,1) (6,1)}; \label{plt:combo_search_k_w_lutnet}
        \addplot [thick, dashed, domain=0.5:6.5,samples=50,variable=\y] ({\y + log2(\y-1) - 1}, {\y}); \label{plt:combo_search_k_w_frontier}
        \addplot [white] coordinates {(2, 4)} node [black, anchor=center] {Infeasible};
    \end{axis}
    
    \newcommand{\inputlen}{2mm}
	\newcommand{\outputlen}{2mm}
	
	\tikzstyle {lut} = [rectangle, minimum width=18mm, minimum height=28mm, align=center, draw]
	\tikzstyle {sum} = [circle, minimum width=7.5mm, minimum height=7.5mm, align=center, draw]
	\tikzstyle {arrow} = [->, >=stealth]
	
	\node (lut1) [lut] at (-0.5,4.5) {};
	\draw ($(lut1.west)+(0,5mm)$) -- ++(left:\inputlen) node [left, anchor=east] {$\tilde{x}^{\left(m,1\right)}_{1}, \tilde{x}^{\left(m,2\right)}_{1}, \cdots$};
	\draw ($(lut1.west)+(0,0)$) -- ++(left:\inputlen) node [left, anchor=east] {$\tilde{x}^{\left(m,1\right)}_{2}, \tilde{x}^{\left(m,2\right)}_{2}, \cdots$};
	\draw ($(lut1.west)+(0,-5mm)$) -- ++(left:\inputlen) node [left, anchor=east] {$\tilde{p}_1^{\left(m,1\right)}, \tilde{p}_1^{\left(m,2\right)}, \cdots$};

	\draw [arrow] (lut1.east) -- ++(right:\outputlen) node [left, anchor=west] {};
	
	\node (word box) [lut, draw=none, scale=0.8] at (lut1.center) {\shortstack{$(c_{(-1,-1,-1)}, $\\$c_{(-1,-1,1)}, $\\$c_{(-1,1,-1)}, $\\$c_{(-1,1,1)}, $\\$c_{(1,-1,-1)}, $\\$c_{(1,1,-1)}, $\\$c_{(1,-1,1)}, $\\$c_{(1,1,1)})$}};
	
	\node (eqn) [right=of lut1.center, xshift=-1mm] {$\equiv$};
	
	\node (lut2) [lut, right=of lut1, xshift=24mm, minimum height=37mm, minimum width=26mm] {};
	
	\node (wordbox1) [lut, minimum height=20mm, minimum width=22mm, scale=0.8, above=of lut2.center, yshift=-14mm, xshift=-2mm] {\shortstack{$(c_{(-1,-1,-1)}, $\\$c_{(-1,-1,1)}, $\\$c_{(-1,1,-1)}, $\\$c_{(-1,1,1)})$}};
	\node (wordbox2) [lut, minimum height=20mm, minimum width=22mm, scale=0.8, below=of lut2.center, yshift=16mm, xshift=-2mm] {\shortstack{$(c_{(1,-1,-1)}, $\\$c_{(1,-1,1)}, $\\$c_{(1,1,-1)}, $\\$c_{(1,1,1)})$}};

	\draw ($(wordbox1.west)+(0,2mm)$) -- ++(left:4mm) node [left, anchor=east] {$\tilde{x}^{\left(m,1\right)}_{1}, \tilde{x}^{\left(m,2\right)}_{1}, \cdots$};
	\draw ($(wordbox1.west)+(0,-2mm)$) -- ++(left:4mm) node [left, anchor=east] {$\tilde{x}^{\left(m,1\right)}_{2}, \tilde{x}^{\left(m,2\right)}_{2}, \cdots$};
	
	\draw ($(wordbox2.west)+(0,2mm)$) -- ++(left:0.7mm) -- ($(wordbox1.west)+(-0.7mm,2mm)$) node [circle, fill=black, inner sep=0.5pt] {};
	\draw ($(wordbox2.west)+(0,-2mm)$) -- ++(left:1.3mm) -- ($(wordbox1.west)+(-1.3mm,-2mm)$) node [circle, fill=black, inner sep=0.5pt] {};
	
	\node (mux1) [right=of lut2, xshift=-16mm] {};
	\draw [fill=none] ([xshift=-1mm, yshift=-4mm]mux1.center) -- ([xshift=-1mm, yshift=4mm]mux1.center) -- ([xshift=1mm, yshift=2mm]mux1.center) -- ([xshift=1mm, yshift=-2mm]mux1.center) -- cycle;
	
	\draw (wordbox1.east) -- ++(right:0.7mm) |- ([xshift=-1mm, yshift=2mm]mux1.center);
	\draw (wordbox2.east) -- ++(right:0.7mm) |- ([xshift=-1mm, yshift=-2mm]mux1.center);

	\draw [arrow] ($(mux1.center)+(1mm,0)$) -- ++(right:3mm) node [left, anchor=west] {};
	
	\draw ($(mux1.center)+(0,-3mm)$) -- ++(down:10.5mm) -- ++(left:21mm) node [left, anchor=east] {$\tilde{p}_1^{\left(m,1\right)}, \tilde{p}_1^{\left(m,2\right)}, \cdots$};
	
	
	\draw [line width=0.10pt] ($(11mm,10mm)$) -- ($(lut1.center)+(-25mm,-15mm)$);
	\draw [line width=0.10pt] ($(14mm,10mm)$) -- ($(lut2.center)+(10mm,-15mm)$);
	\node [circle, draw=black, inner sep=2pt] at (12.4mm,9.75mm) {};

\end{tikzpicture}
			\caption{
				Feasible choices of $K$ and $P$ values for $\left(K, P\right)$-LUTNet, with a demonstration of a $\left(3, 1\right)$-LUT microarchitecture implemented using a 3-LUT.
				The dashed line represents the frontier of feasible microarchitectures.
				We require those with $P = 0$ (\ref{plt:combo_search_k_w_lutnet}) to be fully unrolled, while we allow those with $P > 0$ (\ref{plt:combo_search_k_w_kwlutnet}) to be tiled across both input and output channels.
			}
			\label{plot:search_k_w}
		\end{figure}
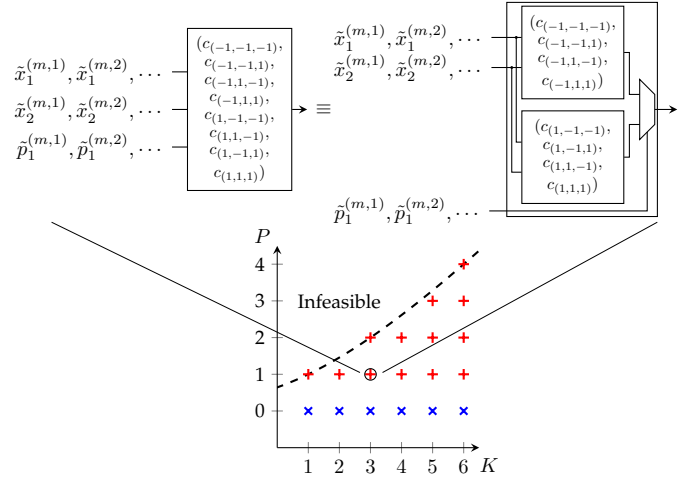

        Each operator $g_m$ produced by our training regime will be mapped to our FPGA-specific microarchitecture, which we denote as $\left(K, P\right)$-LUTNet.
        While ``$\left(K, P\right)$-LUTs" are $K$-LUTs, each can more intuitively be viewed as $2^P$ internal $K - P$-LUTs, sharing all inputs, multiplexed between using $P$ selection bits.
        Fig.~\ref{plot:search_k_w} shows an example with $K = 3$ and $P = 1$.
        Here, two input ports are used for feeding input activations $\tilde{\boldsymbol{x}}$ and one for the multiplexer selection bitstream $\tilde{p}$, which comes from RAM.
        The eight $c$ parameters form the configuration mask of the 3-LUT.
        While each arithmetic operation is associated with a unique $\tilde{p}$, $\boldsymbol{c}$ is shared across all activations computed with the same operator.
        
		Fig.~\ref{plot:search_k_w} also shows all feasible combinations of $K$ and $P$ for $K \leq 6$, the number of inputs per LUT in major vendors' current-generation FPGAs.
        We consider an architecture to be feasible only if the total number of expressible $K - P$-input functions is at least equal to the number of functions that can be selected via the $P$ selection bits, \emph{i.e.} if $2^{2^{K - P}} \geq 2^P$.
        
        When tiled with $P = 0$, a network constructed with LUTs configured in this way would have significantly lower flexibility than an equivalent BNN, resulting in a low classification accuracy.
        We therefore only consider the case where $\left(K, 0\right)$-LUTNet implementations are unrolled.
    
\begin{figure}
    	    \centering
    	    \begin{tikzpicture}[label distance=2mm,decoration={markings,mark= at position 0.5 with{\node[font=\footnotesize] {/};} },scale=0.8, every node/.style={scale=0.7}]

    \tikzstyle{ItemBox} = [rectangle, draw, text centered, minimum height=1em, minimum width=10em, node distance=3cm, text width=9em]
    \tikzstyle{ActionCircle} = [rounded rectangle, draw, text centered, minimum height=1em, minimum width=10em, node distance=3cm, text width=9em]
    \tikzstyle{TextBox} = [rectangle, text centered, minimum height=1em, minimum width=10em, node distance=3cm, text width=9em]
    
    \node at (-1.7,6){}; 
    
    \node[ItemBox] at (0,6) (model) {Model, dataset, activation precision};
    \node[ItemBox] at ($(model)+(0,-0.8)$) (pruning_threshold) {$\theta$};
    \node[ItemBox] at ($(pruning_threshold)+(0,-0.8)$) (tiling_factor) {$K$, $P$, $T$, target layers};
    
    \node[ActionCircle] at ($(model)+(4,0)$) (training) {Training};
    \node[ActionCircle] at ($(pruning_threshold)+(4,0)$) (pruning) {Pruning};
    \node[ActionCircle] at ($(tiling_factor)+(4,0)$) (logic_expansion) {Logic expansion};
    
    \node[ItemBox] at ($(logic_expansion)+(0,-1)$) (trained_lutnet) {Trained network};
    
    \node[ActionCircle] at ($(trained_lutnet)+(-3,-1.5)$) (hls) {High-level synthesis};
    \node[ActionCircle] at ($(trained_lutnet)+(3,-1.5)$) (lut_array_rtl_generation) {LUT array generation};
    
    \node[ItemBox] at ($(trained_lutnet)+(0,-2.5)$) (rtl) {Verilog};
    
    \node[ActionCircle] at ($(rtl)+(0,-1.5)$) (logic_synthesis) {Compilation};
    
    \node[ItemBox] at ($(logic_synthesis)+(0,-1)$) (bitstream) {Bitstream};
    
    \draw [->] (model.east) -- (training.west);
    \draw [->] (pruning_threshold.east) -- (pruning.west);
    \draw [->] (tiling_factor.east) -- (logic_expansion.west);
    \draw [->] (training.south) -- (pruning.north);
    \draw [->] (pruning.south) -- (logic_expansion.north);
    \draw [->] (logic_expansion.south) -- (trained_lutnet.north);
    
    
    \draw [->] (trained_lutnet.west) -| (hls.north);
    \draw [->] (trained_lutnet.east) -| (lut_array_rtl_generation.north);
    \draw [->] (hls.south) |- (rtl.west);
    \draw [->] (lut_array_rtl_generation.south) |- (rtl.east);
    
    \draw [->] (rtl.south) -- (logic_synthesis.north);
    \draw [->] (logic_synthesis.south) -- (bitstream.north);
    
    \draw [dashed] ($(training.north)+(-2,0.8)$) -- ($(training.north)+(2,0.8)$) -- ($(logic_expansion.south)+(2,-0.2)$) -- ($(logic_expansion.south)+(-2,-0.2)$) -- ($(training.north)+(-2,0.8)$);
    
    \draw [dashed] ($(hls.north)+(-2,0.7)$) -- ($(hls.north)+(2,0.7)$) -- ($(hls.south)+(2,-0.2)$) -- ($(hls.south)+(-2,-0.2)$) -- ($(hls.north)+(-2,0.7)$);
    \draw [dashed] ($(lut_array_rtl_generation.north)+(-2,0.7)$) -- ($(lut_array_rtl_generation.north)+(2,0.7)$) -- ($(lut_array_rtl_generation.south)+(2,-0.2)$) -- ($(lut_array_rtl_generation.south)+(-2,-0.2)$) -- ($(lut_array_rtl_generation.north)+(-2,0.7)$);
    \draw [dashed] ($(logic_synthesis.north)+(-2,0.7)$) -- ($(logic_synthesis.north)+(2,0.7)$) -- ($(logic_synthesis.south)+(2,-0.2)$) -- ($(logic_synthesis.south)+(-2,-0.2)$) -- ($(logic_synthesis.north)+(-2,0.7)$);
    
    \node[TextBox] at ($(training.north)+(0,0.4)$) {TensorFlow};
    \node[TextBox] at ($(hls.north)+(-1,0.3)$) {Vivado HLS};
    \node[TextBox] at ($(lut_array_rtl_generation.north)+(1,0.3)$) {Python};
    \node[TextBox] at ($(logic_synthesis.north)+(-1,0.4)$) {Vivado};

\end{tikzpicture}
        	\caption{LUTNet's fully automated training and FPGA implementation flow.}
        	\label{plot:OVERALL_FLOW}
        \end{figure}
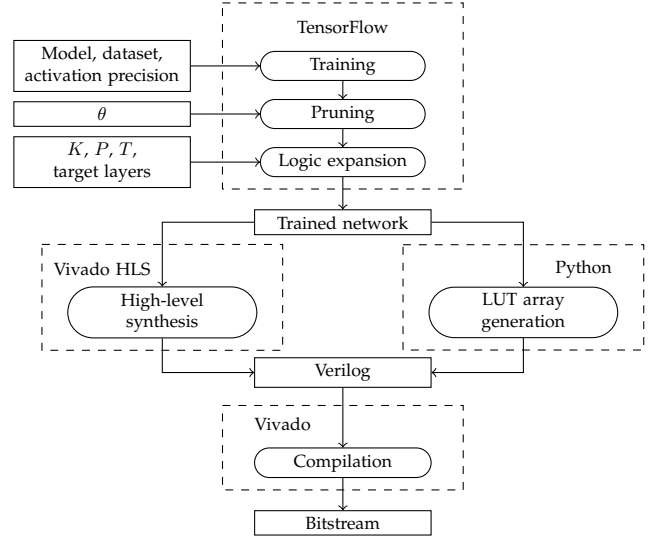

        A representation of the fully automated LUTNet software training and hardware implementation flow is shown in Fig.~\ref{plot:OVERALL_FLOW}.
        As input, the user provides the desired network model, training dataset, activation precision and the required pruning level to our TensorFlow-based training software, which performs training and pruning.
        With user-supplied $K$, $P$, and $T$, logic expansion is then performed on the specified layers to construct the LUTNet architecture.
        
        \begin{table*}
        		\centering
        		\caption{
        		    Network architectures for evaluated benchmarks.
        		    Conv\textsubscript{$x,y,z$} denotes a convolutional layer with $x$ outputs, kernel size $y \times y$, and stride $z$.
        		    FConn\textsubscript{$x$} is a fully connected layer with $x$ outputs.
        		    MaxPool\textsubscript{$x$} is an $x \times x$ maximum-pooling layer, and BatchNorm and SoftMax are batch normalization and normalized exponential layers, respectively.
					For the experiments described in Sections~\ref{sec:eval_area}--\ref{sec:eval_energy}, only the layers shown in bold were implemented following the LUTNet approach; the remainder used the ReBNet architecture.
					In Section~\ref{sec:eval_throughput}, LUTNet was used for all Conv and FConn layers.
                }
    		    \begin{tabular}{ccc}
					\toprule
					Dataset			 		& Model					& Network architecture    																																															\\
					\midrule
					\multirow{2}{*}{MNIST}  & \multirow{2}{*}{LFC}  & FConn\textsubscript{256}, BatchNorm, \textbf{FConn\textsubscript{256}}, BatchNorm, \textbf{FConn\textsubscript{256}}, BatchNorm, \textbf{FConn\textsubscript{256}}, BatchNorm, \textbf{FConn\textsubscript{10}}, 	\\
											&						& BatchNorm, SoftMax 																																																\\
					\midrule
											&   					& Conv\textsubscript{64, 3, 1}, BatchNorm, Conv\textsubscript{64, 3, 1}, BatchNorm, MaxPool\textsubscript{2}, Conv\textsubscript{128, 3, 1}, BatchNorm,	Conv\textsubscript{128, 3, 1}, 								\\
					SVHN \& CIFAR-10		& CNV   				& BatchNorm, MaxPool\textsubscript{2}, Conv\textsubscript{256, 3, 1}, BatchNorm, \textbf{Conv\textsubscript{256, 3, 1}}, BatchNorm, FConn\textsubscript{512}, BatchNorm,  											\\
											&						& FConn\textsubscript{512}, BatchNorm, FConn\textsubscript{10}, BatchNorm, SoftMax																																	\\
					\midrule
											&						& Conv\textsubscript{96, 11, 4}, BatchNorm, MaxPool\textsubscript{3}, Conv\textsubscript{256, 5, 1}, BatchNorm, MaxPool\textsubscript{3}, Conv\textsubscript{384, 3, 1}, BatchNorm, 								\\
					ImageNet				& AlexNet   			& Conv\textsubscript{384, 3, 1}, BatchNorm, \textbf{Conv\textsubscript{256, 3, 1}}, BatchNorm, MaxPool\textsubscript{3}, FConn\textsubscript{4096}, BatchNorm, FConn\textsubscript{4096}, 							\\
											&						& BatchNorm, FConn\textsubscript{1000}, BatchNorm, SoftMax   																																						\\
					\bottomrule
				\end{tabular}
    			\label{tab:model_info}
    		\end{table*}
        
        We chose to target Xilinx devices for this work, for which two parallel synthesis flows are required in order to convert the trained network into Verilog.
        For ease of design and modification, all hardware apart from the inference $K$-LUTs is generated from C templates with Vivado HLS.
        The latter components are assembled into dataflow engines similar in structure to those proposed by Venieris \emph{et al.}~\cite{NR_CNN_FXP_FPGACONVNET} and Umuroglu \emph{et al.}~\cite{BNN_CNN_FINN}.
        The creation of LUT arrays---hardware realizations of the inference node functions learned during training---is outsourced to a custom Verilog generator written in Python, the output of which is substituted for placeholders present in the source previously compiled by Vivado HLS.
        Vivado is then used for final synthesis, implementation, and bitstream generation.
        While our flow currently targets Xilinx parts, LUTNet's principles are applicable to all fine-grained LUT-based FPGAs.
        
        A separate LUT array generator is required because, as a general-purpose C-to-Verilog synthesis tool, Vivado HLS compulsorily performs code transformations and optimizations for the synthesis of efficient hardware.
        Given that LUT configurations are already learned during training, it is unnecessary---and extremely time-consuming---for such optimization to be performed on this logic at the C level. 
        Optimization of large LUT arrays at the netlist level during synthesis with Vivado is a lot more efficient, typically taking a few hours---rather than days or weeks---to complete.
        
	\section{Evaluation}
	
	    \subsection{Benchmarks}
            
            For evaluation, we implemented end-to-end dataflow engines for the DNN models shown in Table~\ref{tab:model_info}, using them to classify the listed datasets.
			All hardware implementations targeted the Xilinx Kintex UltraScale XCKU115 FPGA and met timing at 200~MHz.
            Our baseline was the state-of-the-art BNN architecture, ReBNet.
            For fairness of comparison between LUTNet and ReBNet implementations, identical layer-wise tiling factors were always used.
			
    
        \subsection{Training Particulars}
        
\begin{figure}
                \begin{tikzpicture}
    
    \begin{groupplot} [
        scale only axis,
		width=0.85\columnwidth,
		height=0.22\textwidth,
		group style={group size=1 by 2, xlabels at=edge bottom, vertical sep=2em},
		ymin=0,
		xlabel near ticks,
		xlabel={Epoch},
        ylabel near ticks
	]
        
        \nextgroupplot
            
        \addplot [thick, red] table [y=training_error, x=epoch] {data/cifar_training_curve/train.txt}; \label{plt:cifar_training_curve_train}
        \addplot [thick, black!25!green, densely dashed] table [y=training_error, x=epoch] {data/cifar_training_curve/prune.txt}; \label{plt:cifar_training_curve_prune}
        \addplot [thick, blue, loosely dashed] table [y=training_error, x=epoch] {data/cifar_training_curve/retrain.txt}; \label{plt:cifar_training_curve_retrain}
        
        \node [text width=1em, anchor=north] at (axis description cs:0.5, 1) {\subcaption{\label{plot:cifar_training_curve}}};
            
        \nextgroupplot [
            ylabel={Top-1 training error rate (\%)},
            every axis y label/.append style={at=(ticklabel cs:1.1)}
        ]
        
        \addplot [thick, red] table [y=training_error, x=epoch] {data/imagenet_training_curve/train.txt}; \label{plt:imagenet_training_curve_train}
        \addplot [thick, black!25!green, densely dashed] table [y=training_error, x=epoch] {data/imagenet_training_curve/prune.txt}; \label{plt:imagenet_training_curve_prune}
        \addplot [thick, blue, loosely dashed] table [y=training_error, x=epoch] {data/imagenet_training_curve/retrain.txt}; \label{plt:imagenet_training_curve_retrain}
        
        \node [text width=1em, anchor=north] at (axis description cs:0.5, 1) {\subcaption{\label{plot:imagenet_training_curve}}};
    
    \end{groupplot}

\end{tikzpicture}
            	\caption{
            	    Training loss for \subref{plot:cifar_training_curve} the CNV network classifying the CIFAR-10 dataset and \subref{plot:imagenet_training_curve} AlexNet classifying ImageNet during high-precision training~(\ref{plt:cifar_training_curve_train}), high-precision post-pruning retraining~(\ref{plt:cifar_training_curve_prune}), and post-logic expansion retraining with binarized forward propagation~(\ref{plt:cifar_training_curve_retrain}).
                }
            	\label{plot:training_curves}
            \end{figure}
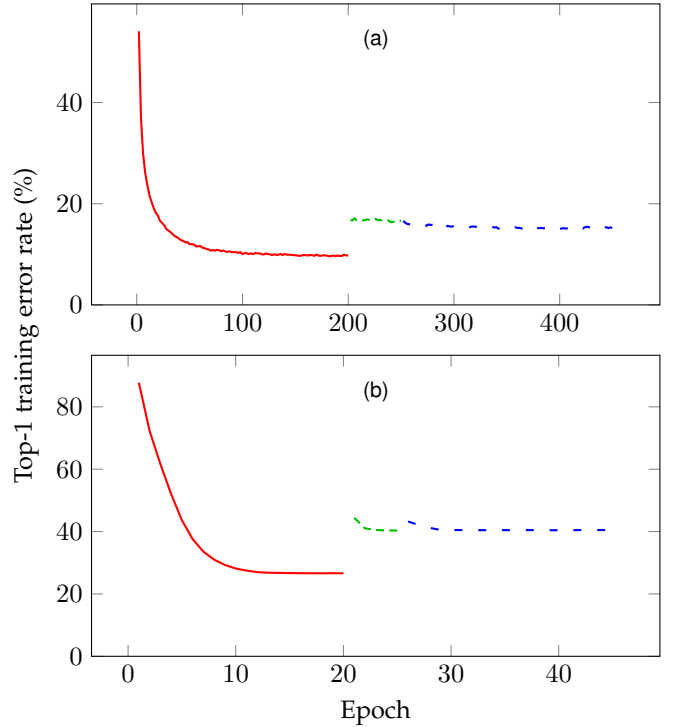

            For our simpler datasets (MNIST~\cite{MNIST}, SVHN, and CIFAR-10), we performed the training, post-pruning retraining, and post-logic expansion retraining described in Section~\ref{sec:training} for 200, 50, and 200 epochs, respectively. 
            For the more complex ImageNet dataset, these were performed for 20, 5, and 20 epochs instead.
            These periods were selected from our observations during training, the loss curves for which are shown in Fig.~\ref{plot:training_curves}, demonstrating saturation at or before these epochs.
            Non-LUTNet implementations were identically trained, but the logic expansion phase (Section~\ref{sec:training_expansion}) was not performed.
            Training phases were executed in TensorFlow and accelerated using four Nvidia GTX~1080~Ti graphics processing units (GPUs).
            
		\subsection{Metrics}
		\label{sec:metrics}
		
		    When evaluating our designs, we were primarily interested in \emph{logic density}, which we define as the number of LUTs required to construct a network able to achieve a particular test accuracy for a given dataset.
			The fewer LUTs needed to reach the same accuracy, the higher the density and thus the more efficient the implementation.
			We evaluated logic density along with secondary metrics energy efficiency, throughput, and training time.
			Logic expansion results in improvements in all but the latter of these, leading to a more favorable multi-dimensional Pareto frontier from which to select design points than available with XNOR-based BNNs.
		
		\subsection{Area Efficiency}
		\label{sec:eval_area}
		
			\subsubsection{Unrolled Implementation}
			\label{sec:eval_area_unrolled}
				
				In order to demonstrate the full capabilities of specialized LUTs, we initially unrolled a subset of each network such that each node within that subset was mapped to a distinct compute unit.
				We unrolled as many layers as the target device could accommodate and implemented them following the $\left(K,0\right)$-LUTNet approach, leaving all other layers unchanged.
				Those selected for unrolled LUTNet implementation are marked in bold in Table~\ref{tab:model_info}.
				For fairness of comparison, the ReBNet baselines had the same layers unrolled, and fine-grained pruning was performed identically to that carried out for LUTNet on those layers.
				The remaining layers were left tiled, with identical tiling factors to those used for ReBNet's evaluation.
				
				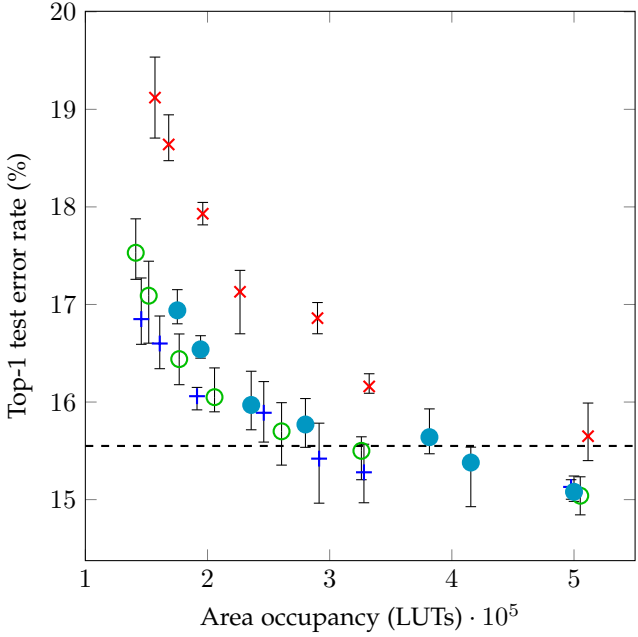
\begin{figure}
					\centering
					\begin{tikzpicture}

    \begin{axis}[
		width=\columnwidth,
		height=\columnwidth,
		xlabel near ticks,
		ylabel near ticks,
		ylabel={Top-1 test error rate (\%)},
        xmin=100000,
        xmax=550000,
        error bars/y dir=both,
        error bars/y explicit=true,
        xtick scale label code/.code={\pgfmathparse{int(#1)}$\text{Area occupancy (LUTs)} \cdot 10^{\pgfmathresult}$},
		every x tick scale label/.style={at={(xticklabel cs:0.5)}, anchor=north}
    ]
        \addplot [thick, only marks, mark=x, mark options={scale=1.5, color=red}] table [y=RPerr, x=RPLUT, y error plus=RerrbarU, y error minus=RerrbarL] {data/err_area.txt}; \label{plt:tradeoff_rebnet}
        \addplot [thick, only marks, mark=o, mark options={scale=1.5, color=black!25!green}] table [y=2Lerr, x=2LLUT, y error plus=2errbarU, y error minus=2errbarL] {data/err_area.txt}; \label{plt:tradeoff_2lutnet}
        \addplot [thick, only marks, mark=+, mark options={scale=1.5, color=blue}] table [y=4Lerr, x=4LLUT, y error plus=4errbarU, y error minus=4errbarL] {data/err_area.txt}; \label{plt:tradeoff_4lutnet}
        \addplot [thick, only marks, mark=*, mark options={scale=1.5, color=black!25!cyan}] table [y=6Lerr, x=6LLUT, y error plus=6errbarU, y error minus=6errbarL] {data/err_area.txt}; \label{plt:tradeoff_6lutnet}
        \addplot [thick, dashed] coordinates {(\pgfkeysvalueof{/pgfplots/xmin},15.55) (\pgfkeysvalueof{/pgfplots/xmax},15.55)};
	\end{axis}

\end{tikzpicture}
					\caption{
						Whole-network area \emph{vs} CIFAR-10 accuracy tradeoffs for unrolled, pruned ReBNet~\cite{BNN_CNN_REBNET_FCCM}~(\ref{plt:tradeoff_rebnet}), $\left(2,0\right)$-LUTNet~(\ref{plt:tradeoff_2lutnet}), $\left(4,0\right)$-LUTNet~(\ref{plt:tradeoff_4lutnet}), and $\left(6,0\right)$-LUTNet~(\ref{plt:tradeoff_6lutnet}) implementations of CNV's sixth convolutional layer.
						All other layers were tiled, unpruned ReBNet realizations.
						Each point is representative of a distinct pruning threshold.
						Error bars show the minimum, mean, and maximum accuracy achieved over five independent training runs.
						The dashed line shows the baseline accuracy for an unrolled, unpruned ReBNet implementation (660196 LUTs).
					}
					\label{plot:AREA_LUT_TRADEOFF}
				\end{figure}
				
				Fig.~\ref{plot:AREA_LUT_TRADEOFF} shows the achieved whole-network area \emph{vs} test accuracy points for ReBNet and $\left(K,0\right)$-LUTNet implementations, each pruned to various densities (proportion of remaining pre-pruning parameters) via the tuning of $\theta$, for CNV classifying the CIFAR-10 dataset.
				Each point marks the mean of five independent training runs, with an error bar indicating maxima and minima.
				LUTNet implementations used 2-, 4-, and 6-LUT inference operators.
				For reference, the mean top-1 test error rate of ReBNet without pruning---again averaged over five training runs---is also shown.
				From these data, one can clearly observe that while the error rate increases as more aggressive pruning is applied, LUTNet demonstrates greater robustness to that pruning than ReBNet through its increased logic density.
				That several LUTNet points achieve greater test accuracy than the unpruned baseline speaks to LUTNet's increased expressiveness.
				For example, despite having a significantly lower (2.27$\times$) area requirement, our 91.1\%-pruned $\left(4,0\right)$-LUTNet implementation achieved an accuracy 0.590 percentage points (pp) above that of the ReBNet implementation without pruning.
				
				
				It is interesting to note from Fig.~\ref{plot:AREA_LUT_TRADEOFF} that $\left(6,0\right)$-LUTNet implementations tended to achieve lower logic densities than those of $\left(4,0\right)$- and sometimes even $\left(2,0\right)$-LUTNet.
				To explain this, we must consider area and accuracy separately.
				
\begin{figure}
					\begin{tikzpicture}
    
    \begin{groupplot} [
		ybar,
		scale only axis,
		width=0.85\columnwidth,
		height=0.22\textwidth,
		group style={group size=1 by 2, xlabels at=edge bottom, xticklabels at=edge bottom, vertical sep=2em},
		xlabel near ticks,
		xlabel={Post-pruning density},
		ylabel absolute,
		every axis y label/.append style={yshift=-1em},
		xtick={1, 2},
        xticklabels={4.02\%, 11.3\%},
        enlarge x limits=0.5,
        legend image code/.code={
            \draw[#1, bar width=6pt, yshift=-0.3em] plot coordinates {(0cm,0.8em)};
        }
	]
        
        \nextgroupplot [
    		ylabel={Top-1 test error rate (\%)}
        ]
            
        \addplot[thick, pattern=vertical lines, pattern color=red] coordinates {(1, 17.83) (2, 16.29)};
        \label{plt:trn_err_rebnet_low}
        \addplot[thick, pattern=north west lines, pattern color=black!25!green] coordinates {(1, 16.42) (2, 15.63)}
        node [pos=0, xshift=-0.6*\pgfkeysvalueof{/pgf/bar width}, rotate=90, anchor=west] {1.41~pp}
        node [pos=1, xshift=-0.6*\pgfkeysvalueof{/pgf/bar width}, rotate=90, anchor=west] {0.660~pp};
        \label{plt:trn_err_2lutnet_low}
        \addplot[thick, pattern=horizontal lines, pattern color=blue] coordinates {(1, 15.92) (2, 15.45)}
        node [pos=0, xshift=0.6*\pgfkeysvalueof{/pgf/bar width}, rotate=90, anchor=west] {1.91~pp}
        node [pos=1, xshift=0.6*\pgfkeysvalueof{/pgf/bar width}, rotate=90, anchor=west] {0.840~pp};
        \label{plt:trn_err_4lutnet_low}
        \addplot[thick, pattern=north east lines, pattern color=black!25!cyan] coordinates {(1, 15.83) (2, 15.54)}
        node [pos=0, xshift=1.8*\pgfkeysvalueof{/pgf/bar width}, rotate=90, anchor=west] {2.00~pp}
        node [pos=1, xshift=1.8*\pgfkeysvalueof{/pgf/bar width}, rotate=90, anchor=west] {0.750~pp};
        \label{plt:trn_err_6lutnet_low}
            
        \node [text width=1em, anchor=north] at (axis description cs:0.5, 1) {\subcaption{\label{plot:TRAINING_LOSS}}};
        
        \nextgroupplot [
            ymin=0,
            ymax= 650000,
    		ytick scale label code/.code={\pgfmathparse{int(#1)}$\text{Area occupancy (LUTs)} \cdot 10^{\pgfmathresult}$},
		    every y tick scale label/.style={at={(yticklabel cs:0.5)}, rotate=90, yshift=1.5em}
        ]
        
        \addplot[thick, pattern=vertical lines, pattern color=red] coordinates {(1, 196117) (2, 332339)};
        \label{plt:util_rebnet}
        \addplot[thick, pattern=north west lines, pattern color=black!25!green] coordinates {(1, 176833) (2, 325959)}
        node [pos=0, xshift=-0.6*\pgfkeysvalueof{/pgf/bar width}, rotate=90, anchor=west] {1.11$\times$}
        node [pos=1, xshift=-0.6*\pgfkeysvalueof{/pgf/bar width}, rotate=90, anchor=west] {1.02$\times$};
        \label{plt:util_2lutnet}
        \addplot[thick, pattern=horizontal lines, pattern color=blue] coordinates {(1, 191405) (2, 328041)}
        node [pos=0, xshift=0.6*\pgfkeysvalueof{/pgf/bar width}, rotate=90, anchor=west] {1.02$\times$}
        node [pos=1, xshift=0.6*\pgfkeysvalueof{/pgf/bar width}, rotate=90, anchor=west] {1.01$\times$};
        \label{plt:util_4lutnet}
        \addplot[thick, pattern=north east lines, pattern color=black!25!cyan] coordinates {(1, 235757) (2, 415529)}
        node [pos=0, xshift=1.8*\pgfkeysvalueof{/pgf/bar width}, rotate=90, anchor=west] {0.832$\times$}
        node [pos=1, xshift=1.8*\pgfkeysvalueof{/pgf/bar width}, rotate=90, anchor=west] {0.800$\times$};
        \label{plt:util_6lutnet}
            
        \node [text width=1em, anchor=north] at (axis description cs:0.5, 1) {\subcaption{\label{plot:UTIL_AT_SAME_PRUNING}}};
    
    \end{groupplot}

\end{tikzpicture}
	
					\caption{
						\subref{plot:TRAINING_LOSS} Accuracy and \subref{plot:UTIL_AT_SAME_PRUNING} area for ReBNet~\cite{BNN_CNN_REBNET_FCCM}~(\ref{plt:trn_err_rebnet_low}), $\left(2,0\right)$-LUTNet~(\ref{plt:trn_err_2lutnet_low}), $\left(4,0\right)$-LUTNet~(\ref{plt:trn_err_4lutnet_low}), and $\left(6,0\right)$-LUTNet~(\ref{plt:trn_err_6lutnet_low}) with CNV, pruned to two densities, classifying CIFAR-10.
						Annotations denote decreases \emph{vs} ReBNet.
					}
					\label{plot:area_accuracy_pruned}
				\end{figure}
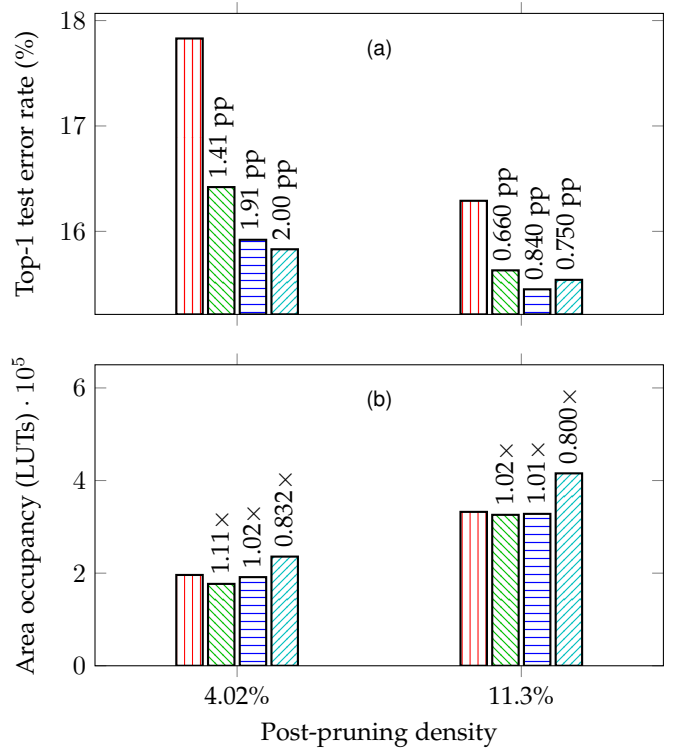

				Fig.~\ref{plot:TRAINING_LOSS} shows the test accuracy of ReBNet, $\left(2,0\right)$-, $\left(4,0\right)$-, and $\left(6,0\right)$-LUTNet pruned to two densities: 4.02\% and 11.3\%.
				These densities were selected for comparison since they represent a wide spread over those found to achieve accuracy reasonably close ($\pm$2.00~pp) to ReBNet when unpruned.
				Of particular pertinence is the difference in accuracy spread between the two: those at 11.3\% are much tighter than their 4.02\% parallels.
				These diminishing accuracy returns when adding LUT inputs at higher densities point to complexity saturation.
			
				Turning now to area, Fig.~\ref{plot:UTIL_AT_SAME_PRUNING} shows the LUT requirements of the same implementations.
				While $(K,0)$-LUTNet designs for any $K$ with equal density contain the same number of logical LUTs, this does not mean that they consume the same number of physical LUTs.
				The LUTs actually present in our target device are 6-LUTs, each capable of implementing either a single logical 6-LUT or two logical $K$-LUTs with at least five (for 5-LUTs), three (4-LUTs) or one (3-LUTs) shared inputs.
				1- and 2-LUTs are not required to share any inputs; two of these can always be packed together.
				For $\left(2,0\right)$- and $\left(4,0\right)$-LUTNet, in which each inference operator uses fewer than five inputs, Vivado can often (for $\left(4,0\right)$-LUTNet) or always ($\left(2,0\right)$-LUTNet) pack two logical $K$-LUTs into each physical 6-LUT, resulting in high logic density.
				Training-induced simplifications, \emph{e.g.} inputs treated as don't-cares that are removed during synthesis, also lead to higher probabilities of additional packing when smaller logical LUTs are used.
				These optimization phenomena are rarely seen for $\left(6,0\right)$-LUTNet, hence its greater area requirements at equal density.
				
				When moving from 4- to 6-LUTs at the higher density, despite the $>$20\% increase in physical LUTs, no accuracy benefit was obtained.
				Due to this, as was shown in Fig.~\ref{plot:AREA_LUT_TRADEOFF}, $\left(4,0\right)$-LUTNet almost always achieves a better area-accuracy tradeoff than $\left(6,0\right)$-LUTNet.
				
\begin{figure}
                    \centering
                    \begin{tikzpicture}
    
    \begin{axis}[ybar,
        scale only axis,
		width=0.85\columnwidth,
		height=0.22\textwidth,
        ymin=0,
        ytick scale label code/.code={\pgfmathparse{int(#1)}$\text{Area occupancy (LUTs)} \cdot 10^{\pgfmathresult}$},
		every y tick scale label/.style={at={(yticklabel cs:0.5)}, rotate=90, anchor=south},
		xtick={1, 2, 3, 4},
        xticklabels={\shortstack{MNIST\\(LFC)}, \shortstack{SVHN\\(CNV)}, \shortstack{CIFAR-10\\(CNV)}, \shortstack{ImageNet\\(AlexNet)}},
        enlarge x limits=0.3,
        x tick label style={rotate=60, anchor=east},
        xlabel near ticks,
        xlabel={Dataset (network)},
        legend image code/.code={
            \draw[#1, bar width=6pt, yshift=-0.3em] plot coordinates {(0cm,0.8em)};
        }
    ]
        \addplot [thick, pattern=vertical lines, pattern color=red] coordinates {(1, 48102) (2, 379403) (3, 511494) (4, 941768)};
        \label{plt:area_comp_rebnet_wup};
        \addplot [thick, pattern=horizontal lines, pattern color=blue] coordinates {(1, 58192) (2, 154814) (3, 246044) (4, 496106)}
        node [pos=0, xshift=0.6*\pgfkeysvalueof{/pgf/bar width}, rotate=90, anchor=west] {0.827$\times$}
        node [pos=0.333, xshift=0.6*\pgfkeysvalueof{/pgf/bar width}, rotate=90, anchor=west] {2.45$\times$}
        node [pos=0.667, xshift=0.6*\pgfkeysvalueof{/pgf/bar width}, rotate=90, anchor=west] {2.08$\times$}
        node [pos=1, xshift=0.6*\pgfkeysvalueof{/pgf/bar width}, rotate=90, anchor=west] {1.90$\times$};
        \label{plt:area_comp_4lutnet};
    \end{axis}

\end{tikzpicture}
	
                	\caption{
                	    Area occupancy for ReBNet~\cite{BNN_CNN_REBNET_FCCM}~(\ref{plt:area_comp_rebnet_wup}) and $\left(4,0\right)$-LUTNet~(\ref{plt:area_comp_4lutnet}) with various models and datasets.
                	    Via pruning, each implementation's test accuracy was kept within $\pm$0.300~pp of that of the unpruned ReBNet baseline's.
                	    Annotations show the area decrease in each case.
                    }
                	\label{plot:area_compression}
                \end{figure}
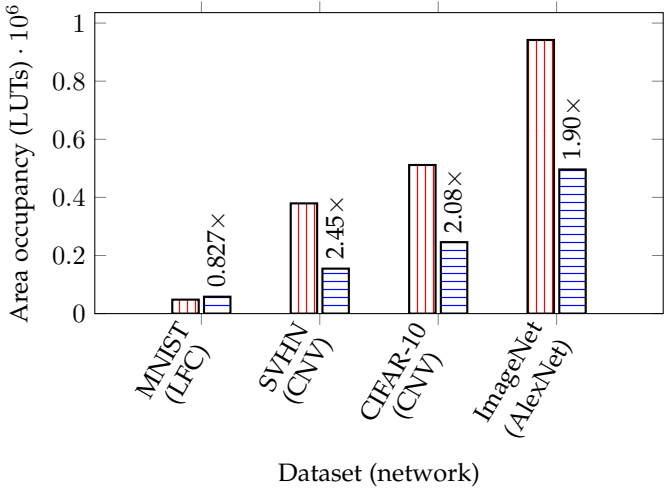

				We also benchmarked LUTNet on other popular datasets and models: MNIST (on LFC), SVHN (on CNV), and ImageNet (on AlexNet).
				Fig.~\ref{plot:area_compression} shows the LUT requirements of each of these model-dataset combinations when implemented using both the ReBNet and $\left(4,0\right)$-LUTNet inference architectures.
				The same layers for all pairs of implementations were unrolled and pruned, with the pruning threshold tuned to achieve an accuracy degradation no more than $\pm$0.300~pp \emph{vs} ReBNet's without pruning.
				
				For CNV and AlexNet, our use of arbitrary inference operators sees area reductions of around 2$\times$.
				For the classification of SVHN, the CNV network used can be pruned more heavily than for CIFAR-10, hence the greater area saving for that dataset.
				For LFC classifying MNIST, however, more LUTs were consumed by $\left(4,0\right)$-LUTNet than its pruned ReBNet counterpart. 
				While each of CNV's hidden layers has 2304 inputs per channel, LFC's channels each have only 256 inputs, presenting less opportunity for area reduction through popcount simplification.
				In this case, $\left(4,0\right)$-LUTNet's area savings through popcount tree thinning were unable to compensate for the inference node LUT incursion.
		
			\subsubsection{Tiled Implementation}
			\label{sec:eval_area_tiled}
			
				Although we have shown that implementing just one network layer using the unrolled LUTNet architecture leads to significant area efficiency gains for a given modern DNN, their complexities make whole-network unrolled LUTNet implementation infeasible.
				Given a fixed-sized FPGA, tiling allows us to trade off throughput and efficiency for additional accuracy by enabling our architecture to be used to implement a greater proportion---including all---of the target network.
                Since tiling can be performed across both input and output channels, to facilitate their distinction we break tiling factor $T$ into two dimensions whose sizes are denoted $T_\text{i}$ and $T_\text{o}$ for input- and output-wise tiling, respectively.
                The overall tiling factor $T = T_\text{i} T_\text{o}$.
                
\begin{figure*}
                    \centering
                    \begin{tikzpicture}
    
    \begin{groupplot}[
        scale only axis,
		width=0.22\textwidth,
		height=0.22\textwidth,
		group style={group size=4 by 1, ylabels at=edge left, yticklabels at=edge left, horizontal sep=1em},
        ylabel near ticks,
        ylabel={Top-1 test error rate (\%)},
        error bars/y dir=both,
        error bars/y explicit=true,
        xtick scale label code/.code={},
        xmin=80000,
        xmax=138000,
        ymin=13.5,
        ymax=23
	]
        
        \nextgroupplot
        
        \addplot [thick, only marks, mark=x, mark options={scale=1.0, color=red}, opacity=0.8] table [y=RTerr, x=RTLUT, y error plus=RTerrbarU, y error minus=RTerrbarL] {data/err_area_tm_k3.txt}; \label{plt:tradeoff_tm_rebnet}
        \addplot [thick, only marks, mark=+, mark options={scale=1.0, color=blue}, opacity=0.8] table [y=31Lerr, x=31LLUT, y error plus=31errbarU, y error minus=31errbarL] {data/err_area_tm_k3.txt}; \label{plt:tradeoff_tm_k3p1_lutnet}
        \addplot [thick, only marks, mark=o, mark options={scale=1.0, color=black!25!green}, opacity=0.8] table [y=32Lerr, x=32LLUT, y error plus=32errbarU, y error minus=32errbarL] {data/err_area_tm_k3.txt}; \label{plt:tradeoff_tm_k3p2_lutnet}
        \addplot [thick, dashed] coordinates {(\pgfkeysvalueof{/pgfplots/xmin},15.55) (\pgfkeysvalueof{/pgfplots/xmax},15.55)};
                
        \node [text width=1em, anchor=north] at (axis description cs:0.5, 1) {\subcaption{\label{plot:AREA_LUT_TRADEOFF_k3}}};
            
        \nextgroupplot [
		    xtick scale label code/.code={\pgfmathparse{int(#1)}$\text{Area occupancy (LUTs)} \cdot 10^{\pgfmathresult}$},
		    every x tick scale label/.style={at={(xticklabel cs:1.05)}, anchor=north}
        ]
        
        \addplot [thick, only marks, mark=x, mark options={scale=1.0, color=red}, opacity=0.8] table [y=RTerr, x=RTLUT, y error plus=RTerrbarU, y error minus=RTerrbarL] {data/err_area_tm_k4.txt};
        \addplot [thick, only marks, mark=+, mark options={scale=1.0, color=blue}, opacity=0.8] table [y=41Lerr, x=41LLUT, y error plus=41errbarU, y error minus=41errbarL] {data/err_area_tm_k4.txt}; \label{plt:tradeoff_tm_k4p1_lutnet}
        \addplot [thick, only marks, mark=o, mark options={scale=1.0, color=black!25!green}, opacity=0.8] table [y=42Lerr, x=42LLUT, y error plus=42errbarU, y error minus=42errbarL] {data/err_area_tm_k4.txt}; \label{plt:tradeoff_tm_k4p2_lutnet}
        \addplot [thick, dashed] coordinates {(\pgfkeysvalueof{/pgfplots/xmin},15.55) (\pgfkeysvalueof{/pgfplots/xmax},15.55)};
                
        \node [text width=1em, anchor=north] at (axis description cs:0.5, 1) {\subcaption{\label{plot:AREA_LUT_TRADEOFF_k4}}};
            
        \nextgroupplot
        
        \addplot [thick, only marks, mark=x, mark options={scale=1.0, color=red}, opacity=0.8] table [y=RTerr, x=RTLUT, y error plus=RTerrbarU, y error minus=RTerrbarL] {data/err_area_tm_k5.txt};
        \addplot [thick, only marks, mark=+, mark options={scale=1.0, color=blue}, opacity=0.8] table [y=51Lerr, x=51LLUT, y error plus=51errbarU, y error minus=51errbarL] {data/err_area_tm_k5.txt}; \label{plt:tradeoff_tm_k5p1_lutnet}
        \addplot [thick, only marks, mark=o, mark options={scale=1.0, color=black!25!green}, opacity=0.8] table [y=52Lerr, x=52LLUT, y error plus=52errbarU, y error minus=52errbarL] {data/err_area_tm_k5.txt}; \label{plt:tradeoff_tm_k5p2_lutnet}
        \addplot [thick, only marks, mark=*, mark options={scale=1.0, color=black!25!cyan}, opacity=0.8] table [y=53Lerr, x=53LLUT, y error plus=53errbarU, y error minus=53errbarL] {data/err_area_tm_k5.txt}; \label{plt:tradeoff_tm_k5p3_lutnet}
        \addplot [thick, dashed] coordinates {(\pgfkeysvalueof{/pgfplots/xmin},15.55) (\pgfkeysvalueof{/pgfplots/xmax},15.55)};
                
        \node [text width=1em, anchor=north] at (axis description cs:0.5, 1) {\subcaption{\label{plot:AREA_LUT_TRADEOFF_k5}}};
            
        \nextgroupplot
        
        \addplot [thick, only marks, mark=x, mark options={scale=1.0, color=red}, opacity=0.8] table [y=RTerr, x=RTLUT, y error plus=RTerrbarU, y error minus=RTerrbarL] {data/err_area_tm_k6.txt};
        \addplot [thick, only marks, mark=+, mark options={scale=1.0, color=blue}, opacity=0.8] table [y=61Lerr, x=61LLUT, y error plus=61errbarU, y error minus=61errbarL] {data/err_area_tm_k6.txt}; \label{plt:tradeoff_tm_k6p1_lutnet}
        \addplot [thick, only marks, mark=o, mark options={scale=1.0, color=black!25!green}, opacity=0.8] table [y=62Lerr, x=62LLUT, y error plus=62errbarU, y error minus=62errbarL] {data/err_area_tm_k6.txt}; \label{plt:tradeoff_tm_k6p2_lutnet}
        \addplot [thick, only marks, mark=*, mark options={scale=1.0, color=black!25!cyan}, opacity=0.8] table [y=63Lerr, x=63LLUT, y error plus=63errbarU, y error minus=63errbarL] {data/err_area_tm_k6.txt}; \label{plt:tradeoff_tm_k6p3_lutnet}
        \addplot [thick, only marks, mark=diamond, mark options={scale=1.0, color=black!25!yellow}, opacity=0.8] table [y=64Lerr, x=64LLUT, y error plus=64errbarU, y error minus=64errbarL] {data/err_area_tm_k6.txt}; \label{plt:tradeoff_tm_k6p4_lutnet}
        \addplot [thick, dashed] coordinates {(\pgfkeysvalueof{/pgfplots/xmin},15.55) (\pgfkeysvalueof{/pgfplots/xmax},15.55)};
                
        \node [text width=1em, anchor=north] at (axis description cs:0.5, 1) {\subcaption{\label{plot:AREA_LUT_TRADEOFF_k6}}};
    
    \end{groupplot}

\end{tikzpicture}
	
                	\caption{
                	    Area-accuracy tradeoffs for pruned ReBNet~\cite{BNN_CNN_REBNET_FCCM}~(\ref{plt:tradeoff_tm_rebnet}) and LUTNet implementations of CNV classifying CIFAR-10 with tiling factors $T_\text{i} = T_\text{o} = 8$.
                	    Across the subplots, we show results for $\left(K,P\right)$-LUTNet for feasible combinations of $K \in \left\{3,\cdots,6\right\}$ and $P > 0$.
                	    In \subref{plot:AREA_LUT_TRADEOFF_k3}, these are $K = 3$ with $P = 1$~(\ref{plt:tradeoff_tm_k3p1_lutnet}) and $2$~(\ref{plt:tradeoff_tm_k3p2_lutnet}); in \subref{plot:AREA_LUT_TRADEOFF_k4}, $K = 4$ with $P = 1$~(\ref{plt:tradeoff_tm_k4p1_lutnet}) and $2$~(\ref{plt:tradeoff_tm_k4p2_lutnet}); in \subref{plot:AREA_LUT_TRADEOFF_k5}, $K = 5$ with $P = 1$~(\ref{plt:tradeoff_tm_k5p1_lutnet}), $2$~(\ref{plt:tradeoff_tm_k5p2_lutnet}), and $3$~(\ref{plt:tradeoff_tm_k5p3_lutnet}); and, in \subref{plot:AREA_LUT_TRADEOFF_k6}, $K = 6$ with $P = 1$~(\ref{plt:tradeoff_tm_k6p1_lutnet}), $2$~(\ref{plt:tradeoff_tm_k6p2_lutnet}), $3$~(\ref{plt:tradeoff_tm_k6p3_lutnet}), and $4$~(\ref{plt:tradeoff_tm_k6p4_lutnet}).
                	    Each point represents a distinct pruning threshold.
						Error bars show the minimum, mean, and maximum accuracy achieved over five independent training runs.
                	    The dashed lines show the baseline accuracy for unpruned ReBNet (133418 LUTs).
                    }
                	\label{AREA_LUT_TRADEOFF_SEARCH_K4}
                \end{figure*}
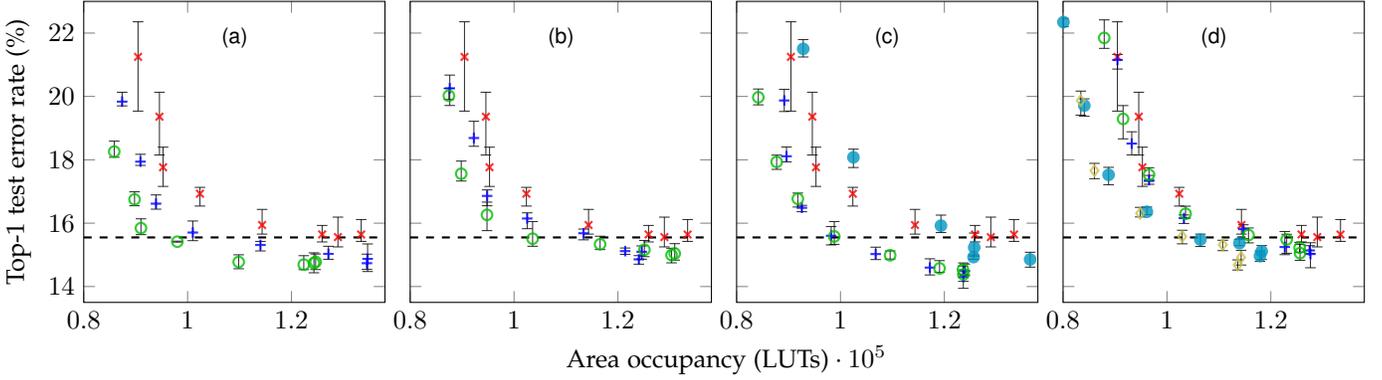

                In order to explore the tradeoffs between $K$, the number of RAM connections per node $P$, area, and accuracy in the presence of tiling, we repeated the experiments performed for Fig.~\ref{plot:AREA_LUT_TRADEOFF} but with the LUTNet layer implemented with $T_\text{i} = T_\text{o} = 8$.
                Fig.~\ref{AREA_LUT_TRADEOFF_SEARCH_K4} shows the results of these for $K \in \left\{3,\cdots,6\right\}$ and all feasible $P$ for each $K$.
                Two conclusions can quickly be drawn from these data.
                Firstly, with the exception of the more heavily pruned $\left(5,3\right)$-LUTNet design points, tiled LUTNet implementations perform favorably compared to their ReBNet counterparts.
                For example, our 64.6\%-pruned $\left(5,1\right)$-LUTNet implementation occupies 1.28$\times$ less area than the equivalently tiled, 16.5\%-pruned ReBNet, with both delivering test accuracy comparable (within $\pm$0.300~pp) to our unpruned ReBNet baseline.
                Secondly, and conversely, the gains in area and accuracy over ReBNet we see with tiling are smaller than those without.
                This is unsurprising; the LUTNet approach becomes less effective when LUTs trained to perform specialized functions are used repeatedly.
				
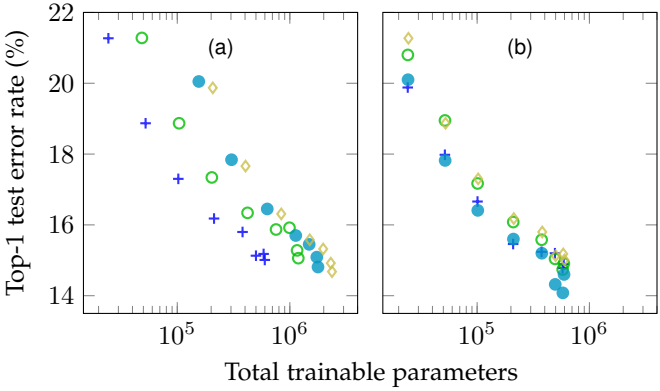
\begin{figure}
                    \centering
                    \begin{tikzpicture}

    \begin{groupplot}[
        scale only axis,
		width=0.41\columnwidth,
		height=0.22\textwidth,
		group style={group size=2 by 1, ylabels at=edge left, yticklabels at=edge left, horizontal sep=1em},
		ylabel near ticks,
		ylabel={Top-1 test error rate (\%)},
        error bars/y dir=both,
        error bars/y explicit=true,
        xmode=log,
        xmax=4000000,
        ymin=13.5,
        ymax=22
    ]
    
        \nextgroupplot[
		    xlabel near ticks,
		    xlabel={Total trainable parameters},
            every axis x label/.append style={at=(ticklabel cs:1.05)}
        ]
        
        \addplot [thick, only marks, mark=+, mark options={scale=1.0, color=blue}, opacity=0.8] table [y=61Lerr, x=61Lparams] {data/params_err_k6.txt}; \label{plt:params_err_k6p1}
        \addplot [thick, only marks, mark=o, mark options={scale=1.0, color=black!25!green}, opacity=0.8] table [y=62Lerr, x=62Lparams] {data/params_err_k6.txt}; \label{plt:params_err_k6p2}
        \addplot [thick, only marks, mark=*, mark options={scale=1.0, color=black!25!cyan}, opacity=0.8] table [y=63Lerr, x=63Lparams] {data/params_err_k6.txt}; \label{plt:params_err_k6p3}
        \addplot [thick, only marks, mark=diamond, mark options={scale=1.0, color=black!25!yellow}, opacity=0.8] table [y=64Lerr, x=64Lparams] {data/params_err_k6.txt}; \label{plt:params_err_k6p4}
        
        \node [text width=1em, anchor=north] at (axis description cs:0.5, 1)
        {\subcaption{\label{plot:params_err_k6}}};
        
        \nextgroupplot
        
        \addplot [thick, only marks, mark=+, mark options={scale=1.0, color=blue}, opacity=0.8] table [y=31Lerr, x=31Lparams] {data/params_err_k3.txt}; \label{plt:params_err_p1k3}
        \addplot [thick, only marks, mark=o, mark options={scale=1.0, color=black!25!green}, opacity=0.8] table [y=41Lerr, x=41Lparams] {data/params_err_k4.txt}; \label{plt:params_err_p1k4}
        \addplot [thick, only marks, mark=*, mark options={scale=1.0, color=black!25!cyan}, opacity=0.8] table [y=51Lerr, x=51Lparams] {data/params_err_k5.txt}; \label{plt:params_err_p1k5}
        \addplot [thick, only marks, mark=diamond, mark options={scale=1.0, color=black!25!yellow}, opacity=0.8] table [y=61Lerr, x=61Lparams] {data/params_err_k6.txt}; \label{plt:params_err_p1k6}
        
        \node [text width=1em, anchor=north] at (axis description cs:0.5, 1)
        {\subcaption{\label{plot:params_err_p1}}};
        
	\end{groupplot}

\end{tikzpicture}
                	\caption{
                        Number of parameters \emph{vs} accuracy for implementations shown in Fig.~\ref{AREA_LUT_TRADEOFF_SEARCH_K4} with fixed \subref{plot:params_err_k6} $K$ and \subref{plot:params_err_p1} $P$.
                        In \subref{plot:params_err_k6}, these feature $K = 6$ with $P = 1$~(\ref{plt:params_err_k6p1}), $2$~(\ref{plt:params_err_k6p2}), $3$~(\ref{plt:params_err_k6p3}), and $4$~(\ref{plt:params_err_k6p4}), while, in \subref{plot:params_err_p1}, we show designs with $P = 1$ and $K = 3$~(\ref{plt:params_err_p1k3}), $4$~(\ref{plt:params_err_p1k4}), $5$~(\ref{plt:params_err_p1k5}), and $6$~(\ref{plt:params_err_p1k6}).
                    }
                	\label{plot:params_err}
                \end{figure}

				Understanding the relative area-accuracy behaviors shown across Fig.~\ref{AREA_LUT_TRADEOFF_SEARCH_K4} requires knowledge of the relationships between $K$, $P$, $T_\text{i}$, $T_\text{o}$, and the total number of trainable parameters.
				All else being equal, tiling \emph{decreases} the total number of trainable LUT parameters by a factor of $T_\text{i} T_\text{o}$ \emph{vs} an unrolled implementation.
				The number of parameters fed from RAM, however, \emph{increases} linearly with $P$, $T_\text{i}$, and $T_\text{o}$.
				With pruning, both quantities will be reduced by a factor of $\nicefrac{1}{\omega}$, where $\omega$ is the post-pruning density.
				Overall, the total number of LUTNet-related parameters within our implementations is $9216\omega{\left(\nicefrac{2^K}{T_\text{i} T_\text{i}} + P T_\text{i} T_\text{o}\right)}$, where 9216 is the number of inference nodes in our unpruned target layer.
				To facilitate analysis, we show these values for a selection of the design points in Fig.~\ref{AREA_LUT_TRADEOFF_SEARCH_K4} in Fig.~\ref{plot:params_err}.
				Note the logarithmic $x$-axis, which we used due to the relative magnitudes of fixed $K$ (Fig.~\ref{plot:params_err_k6}) and $P$ (Fig.~\ref{plot:params_err_p1}) implementations' parameter spaces as well as the decaying exponential relationship between the number of parameters and accuracy.
                
                Comparing between the plots of Fig.~\ref{AREA_LUT_TRADEOFF_SEARCH_K4} reveals a weak trend of worsening area-accuracy behavior with increasing $K$, with curves moving slightly rightwards between Fig.~\ref{plot:AREA_LUT_TRADEOFF_k3} and \subref{plot:AREA_LUT_TRADEOFF_k6}.
                In common with unrolled LUTNet implementations, larger $K$ limits opportunities for double-logical-to-physical LUT packing, decreasing area efficiency.
                These effects tend not to be outweighed by the slight accuracy improvements brought about through the expansion of our parameter space.
                Although the $K$-influenced component of the total number of parameters increases exponentially with $K$, its scaling by $\nicefrac{1}{T_\text{i} T_\text{o}}$---in this case, $\nicefrac{1}{64}$---means that the effects of increasing it are typically small for sensible choices of $K$.
                This is exemplified in Fig.~\ref{plot:params_err_p1} for $P = 1$.
                Relatively, these gaps become even tighter for $P > 1$, since the $P$ component of the parameter space's size is, although only linearly related to $P$, scaled by a much larger factor: $T_\text{i} T_\text{o}$.
                
                The significant increase in total parameters with $P$ is demonstrated in Fig.~\ref{plot:params_err_k6} for $K = 6$.
                With a few exceptions for heavily pruned cases, we see a weak negative correlation between complexity and accuracy here, suggesting that increasing our training space via $P$ is not particularly effective in improving performance.
                These effects tend to be more than outweighed, however, by decreases in area.
                Apart from the anomalous results for $\left(5,3\right)$-LUTNet, for which this relationship unexpectedly reverses, Fig.~\ref{AREA_LUT_TRADEOFF_SEARCH_K4} consistently shows improved area-accuracy tradeoff with increasing $P$.
                We believe that this is due to the structured \emph{vs} unstructured nature of routing connections made from RAM and activations, respectively.
                With fixed $K$, increased $P$ enlarges the ratio of RAM-to-activation connections.
                Since signals from RAM are clustered into buses, while those from the previous network layer tend to be haphazard in structure, the regularity induced through higher $P$ affords Vivado more opportunity for denser LUT packing, lowering area.
                
                Finally, while complexity saturation is apparent for all $\left(K,P\right)$ at higher pruning levels, there is little evidence of overfitting in either Fig.~\ref{AREA_LUT_TRADEOFF_SEARCH_K4} or Fig.~\ref{plot:params_err}; non-negligible downturns in accuracy do not occur with greater numbers of parameters.
                We can therefore conclude that, even with significant complexity increases over ReBNet, LUTNet-based networks are amenable to effective training using standard backward propagation techniques.
		
		\subsection{Area Breakdown}
		
\begin{figure*}
                \centering
                \begin{tikzpicture}
    
    \begin{groupplot}[
        ybar stacked,
        group style={
            columns=2,
            horizontal sep=8em
        },
        axis y line*=left,
        scale only axis,
        height=0.22\textwidth,
        ymin=0,
        xtick=data,
        x tick label style={rotate=60, anchor=east},
        xtick align=outside,
        ytick scale label code/.code={\pgfmathparse{int(#1)}$\text{Area occupancy (LUTs)} \cdot 10^{\pgfmathresult}$},
		every y tick scale label/.style={at={(yticklabel cs:0.5)}, rotate=90, anchor=south},
        legend image code/.code={
            \draw[#1, bar width=6pt, yshift=-0.3em] plot coordinates {(0cm,0.8em)};
        }
    ]
            
        \nextgroupplot [
            width=0.28\textwidth,
		    xticklabels from table={data/util_detail.txt}{Name}
        ]
        
        \addplot [thick, pattern=vertical lines, pattern color=red] table [x=id, y=otherlayers] {data/util_detail.txt};
        \label{plt:breakdown_otherlayers}
        \addplot [thick, pattern=north west lines, pattern color=black!25!green] table [x=id, y=mux] {data/util_detail.txt};
        \label{plt:breakdown_mux}
        \addplot [thick, pattern=horizontal lines, pattern color=blue] table [x=id, y=ops] {data/util_detail.txt};
        \label{plt:breakdown_ops}
        \addplot [thick, pattern=north east lines, pattern color=black!25!cyan] table [x=id, y=popcount] {data/util_detail.txt}
        node [pos=0.2, rotate=90, anchor=west] {1.57$\times$}
        node [pos=0.4, rotate=90, anchor=west] {1.69$\times$}
        node [pos=0.6, rotate=90, anchor=west] {2.08$\times$}
        node [pos=0.8, rotate=90, anchor=west] {1.87$\times$}
        node [pos=1, rotate=90, anchor=west] {2.17$\times$};
        \label{plt:breakdown_popcount}
        
        \node [text width=1em, anchor=north] at (axis description cs:0.5, 1)
        {\subcaption{\label{plot:breakdown_unrolled}}};
        
        \nextgroupplot [
            width=0.46\textwidth,
            ymax=160000,
            xticklabels from table={data/util_detail_tiled.txt}{Name}
        ]
        
        \addplot [thick, pattern=vertical lines, pattern color=red] table [x=id, y=otherlayers] {data/util_detail_tiled.txt};
        \addplot [thick, pattern=north west lines, pattern color=black!25!green] table [x=id, y=mux] {data/util_detail_tiled.txt};
        \addplot [thick, pattern=horizontal lines, pattern color=blue] table [x=id, y=ops] {data/util_detail_tiled.txt};
        \addplot [thick, pattern=north east lines, pattern color=black!25!cyan] table [x=id, y=popcount] {data/util_detail_tiled.txt}
        node [pos=0.111, rotate=90, anchor=west] {1.11$\times$}
        node [pos=0.222, rotate=90, anchor=west] {1.08$\times$}
        node [pos=0.333, rotate=90, anchor=west] {1.28$\times$}
        node [pos=0.444, rotate=90, anchor=west] {1.26$\times$}
        node [pos=0.556, rotate=90, anchor=west] {1.26$\times$}
        node [pos=0.667, rotate=90, anchor=west] {1.10$\times$}
        node [pos=0.778, rotate=90, anchor=west] {1.09$\times$}
        node [pos=0.889, rotate=90, anchor=west] {1.18$\times$}
        node [pos=1, rotate=90, anchor=west] {1.22$\times$};
            
        \node [text width=1em, anchor=north] at (axis description cs:0.5, 1)
        {\subcaption{\label{plot:breakdown_tiled}}};
        
    \end{groupplot}

    \begin{groupplot}[
        group style={
            columns=2,
            horizontal sep=8em
        },
        axis x line=none,
        axis y line*=right,
        scale only axis,
        height=0.22\textwidth,
        ylabel near ticks,
        ylabel={Post-pruning density (\%)}
    ]
            
        \nextgroupplot [
            width=0.28\textwidth
        ]
        
        \addplot [thick, only marks, mark=x, mark options={scale=1.5}] table [x=id, y=Density] {data/util_detail.txt};
        \label{plt:breakdown_density}
        
        \nextgroupplot [
            width=0.46\textwidth
        ]
        
        \addplot [thick, only marks, mark=x, mark options={scale=1.5}] table [x=id, y=Density] {data/util_detail_tiled.txt};
        
    \end{groupplot}

\end{tikzpicture}
            	\caption{
            	    LUT use breakdown, presented in terms of popcount operators~(\ref{plt:breakdown_popcount}), inference operators~(\ref{plt:breakdown_ops}), multiplexing logic~(\ref{plt:breakdown_mux}), and other layers~(\ref{plt:breakdown_otherlayers}), for CNV implementations.
            	    In \subref{plot:breakdown_unrolled}, designs were unrolled; those in \subref{plot:breakdown_tiled} were tiled with $T_\text{i} = T_\text{o} = 8$.
            	    Annotations show decreases \emph{vs} pruned ReBNet.
            	    Each implementation's test accuracy was within $\pm$0.300~pp of that of the unpruned ReBNet baseline's~\cite{BNN_CNN_REBNET_FCCM}.
            	    Points~(\ref{plt:breakdown_density}) show post-pruning densities.
                }
            \label{plot:breakdown}
            \end{figure*}
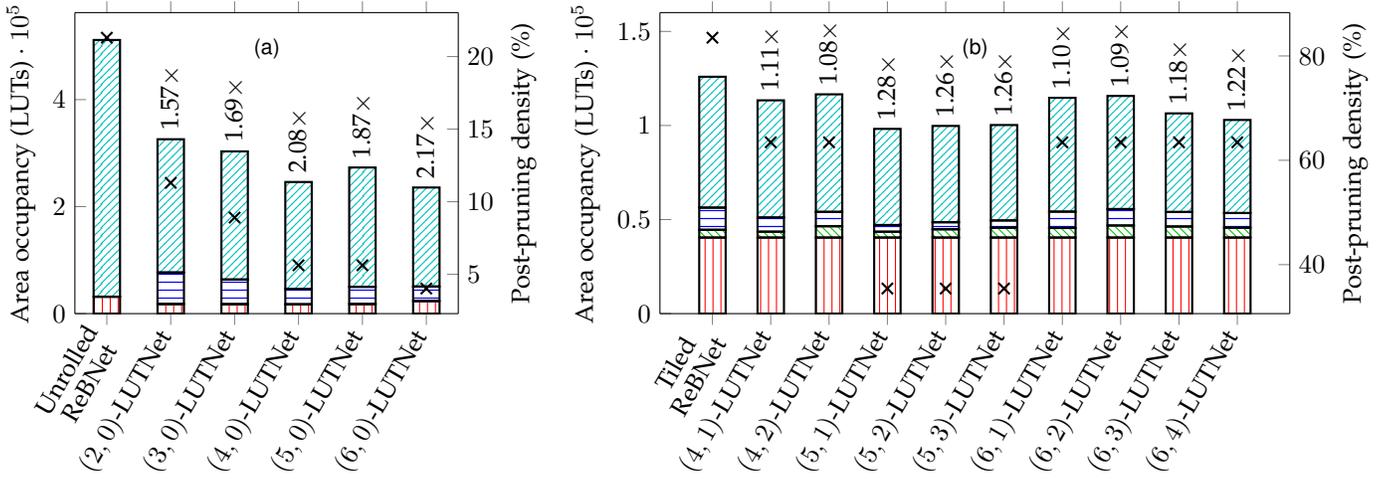

			As a crude method of verifying the source of LUTNet's area savings, we disabled design hierarchy optimization in Vivado, preventing the synthesis engine from flattening across modules.
			By taking a slice of implementations shown in Figs.~\ref{plot:AREA_LUT_TRADEOFF} and \ref{AREA_LUT_TRADEOFF_SEARCH_K4} at the unpruned ReBNet error rate (84.5\%) $\pm$0.300~pp, we obtained ReBNet and LUTNet implementations for CNV all of comparable CIFAR-10 test accuracy.
            Fig.~\ref{plot:breakdown} shows the LUT requirements for each of these, with area split into that required by popcount operators, inference operators, multiplexing logic, and everything else.
            The overall height of each bar is the whole design's area occupancy with hierarchy optimization \emph{enabled}, but the height of each stacked bar is relative to the proportional area obtained with hierarchy optimization \emph{disabled}.
				
            We can see from Fig.~\ref{plot:breakdown_unrolled} that, as more inputs are used per logical LUT, physical LUT requirements generally decrease, highlighting $\left(K,0\right)$-LUTNet's increasing logic density with $K$.
            Each implementation's post-pruning density is also shown.
            From the breakdowns, it can be seen that the number of physical LUTs required for popcount operators drops dramatically with density.
            More aggressive pruning reduces the number of branches in the popcount trees, which consume the majority of the target device's area.
            
            As was pointed out in Section~\ref{sec:intro}, due to following a traditional BNN inference paradigm, unrolled ReBNet implementations---whether pruned or not---require zero LUTs for the realization of their inference operators since their XNOR gates become free-to-implement buffers and inverters.
            For LUTNet, this is not the case: physical LUTs are always consumed by our logical $K$-LUTs.
            As shown in Fig.~\ref{plot:breakdown_unrolled}, however, this is more than outweighed by significant popcount area reduction.
            This confirms the statement made in Section~\ref{sec:intro} regarding $\tilde{N} \ll N$.
            
            Between $\left(2,0\right)$- and $\left(6,0\right)$-LUTNet, we can observe a general trend of decreasing inference operator LUT requirements with density.
            Looking more closely, some more interesting features emerge.
            The jump in total area between $\left(4,0\right)$- and $\left(5,0\right)$-LUTNet can be attributed to two factors: lack of density reduction and decreased opportunity for LUT sharing.
            Here, the increased expressiveness of 5-LUTs was not significant enough to enable increased pruning while remaining within the required accuracy bound.
            On top of this, the logical-to-physical LUT packing effects discussed in Section~\ref{sec:eval_area_unrolled} were marked, pushing both inference operator and total LUT requirements for $\left(5,0\right)$-LUTNet above those for $\left(4,0\right)$-LUTNet.
            Thereafter, although a greater number of physical LUTs were occupied by the $\left(6,0\right)$-LUTNet implementation, a decrease in density facilitated through increased network complexity caused a more-than-compensatory popcount area reduction.
        
            In Fig.~\ref{plot:breakdown_tiled}, we show area information in the same form as Fig.~\ref{plot:breakdown_unrolled} for the tiled LUTNet implementations in Fig.~\ref{AREA_LUT_TRADEOFF_SEARCH_K4} with $K \in \left\{4,5,6\right\}$ and all feasible choices of $P$.
            As expected, tiled implementations' area savings are lower than those for unrolled designs due to their lower tolerance to pruning, although we still see gains over ReBNet in all cases.
                
            It is important to note that, while unrolled LUTNet implementations can achieve significantly greater area savings over ReBNet than their tiled counterparts, they are still physically large.
            For example, while our unrolled $\left(5,0\right)$-LUTNet design was some 1.87$\times$ more compact than unrolled ReBNet due to its 94.4\% sparsity, it was 2.78$\times$ the size of $\left(5,1\right)$-LUTNet with $T_\text{i} T_\text{o} = 64$, which reaches comparable accuracy, albeit at greatly reduced throughput.
            The choice of whether or not to tile---and to what extent if so---depends on the number of LUTs one can afford to use.
            With one-to-one $g_n \to \text{LUT}$ binding, unrolled LUTNet makes the most efficient use of soft logic yet consumes significant numbers of resources, while tiled implementations sacrifice both area efficiency and speed in return for demanding lower area.
		
		\subsection{Implications of Tiling}
		
			While the relationships between tiling factors $T_\text{i}$ and $T_\text{o}$ and area and throughput are straightforward---area and throughput both scale with $\nicefrac{1}{T_\text{i} T_\text{o}}$---those with accuracy are less so.
			Clearly, increased $T_\text{i} T_\text{o}$ will result in accuracy degradation.
			It is not obvious, however, whether tiling across either input or output channels, or some combination of the two, is preferable.
			In Fig.~\ref{plot:robustness}, we show results for the same experiment performed for Fig.~\ref{AREA_LUT_TRADEOFF_SEARCH_K4}, but with $\left(K,P\right)$ fixed at $\left(5,1\right)$ and the constraints on $T_\text{i}$ and $T_\text{o}$ relaxed.
				
			\begin{figure}
				\centering
				\begin{tikzpicture}

    \begin{groupplot}[
        scale only axis,
		width=0.41\columnwidth,
		height=0.22\textwidth,
		group style={group size=2 by 1, ylabels at=edge left, yticklabels at=edge left, horizontal sep=1em},
		ylabel near ticks,
		ylabel={Top-1 test error rate (\%)},
        ymin=13,
        ymax=36,
        error bars/y dir=both,
        error bars/y explicit=true
    ]
    
        \nextgroupplot [
            xlabel near ticks,
            xlabel={Post-pruning density (\%)},
            every axis x label/.append style={at=(ticklabel cs:1.05)}
        ]
        
        \addplot [thick, only marks, mark=x, mark options={scale=1.5, color=red}] table [y=84Lerr, x=84density] {data/err_area_tm_search_t.txt};
        \label{plt:tradeoff_search_t_84}
        \addplot [thick, only marks, mark=+, mark options={scale=1.5, color=blue}] table [y=88Lerr, x=88density] {data/err_area_tm_search_t.txt};
        \label{plt:tradeoff_search_t_88}
        \addplot [thick, only marks, mark=*, mark options={scale=1.5, color=black!25!cyan}] table [y=816Lerr, x=816density] {data/err_area_tm_search_t.txt};
        \label{plt:tradeoff_search_t_816}
        
        \node [text width=1em, anchor=north] at (axis description cs:0.5, 1)
        {\subcaption{\label{plot:robustness_ti8}}};
        
        \nextgroupplot
        
        \addplot [thick, only marks, mark=o, mark options={scale=1.5, color=black!25!green}] table [y=48Lerr, x=48density] {data/err_area_tm_search_t.txt};
        \label{plt:tradeoff_search_t_48}
        \addplot [thick, only marks, mark=+, mark options={scale=1.5, color=blue}] table [y=88Lerr, x=88density] {data/err_area_tm_search_t.txt};
        \addplot [thick, only marks, mark=diamond, mark options={scale=1.5, color=black!25!yellow}] table [y=168Lerr, x=168density] {data/err_area_tm_search_t.txt};
        \label{plt:tradeoff_search_t_168}
        
        \node [text width=1em, anchor=north] at (axis description cs:0.5, 1)
        {\subcaption{\label{plot:robustness_to8}}};
        
	\end{groupplot}

\end{tikzpicture}
				\caption{
				    Density \emph{vs} CIFAR-10 accuracy for tiled $\left(5,1\right)$-LUTNet CNV implementations.
				    In \subref{plot:robustness_ti8}, we show designs with $T_\text{i} = 8$ and $T_\text{o} = 4$~(\ref{plt:tradeoff_search_t_84}), $8$~(\ref{plt:tradeoff_search_t_88}), and $16$~(\ref{plt:tradeoff_search_t_816}), while \subref{plot:robustness_to8} features $T_\text{o} = 8$ and $T_\text{i} = 4$~(\ref{plt:tradeoff_search_t_48}), $8$~(\ref{plt:tradeoff_search_t_88}), and $16$~(\ref{plt:tradeoff_search_t_168}).
				    Each point represents a distinct pruning threshold.
				}
			    \label{plot:robustness}
			\end{figure}
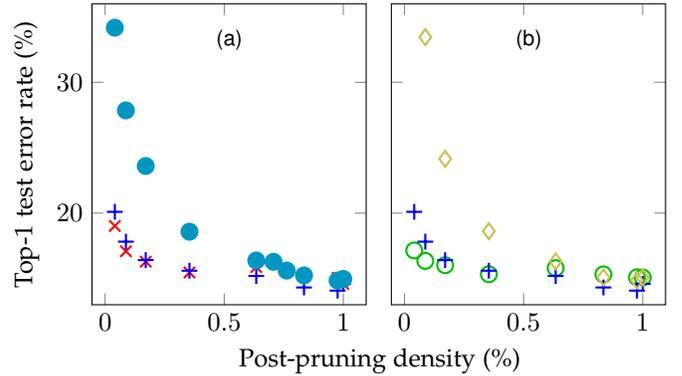
			
			\begin{figure*}
                \centering
                \begin{tikzpicture}
    
    \begin{groupplot}[
        ybar stacked,
        group style={
            columns=2,
            ylabels at=edge left,
            horizontal sep=2em
        },
        scale only axis,
        height=0.22\textwidth,
        ymin=0,
        xtick=data,
        x tick label style={rotate=60, anchor=east},
        xtick align=outside,
        ylabel near ticks,
        ylabel={Power consumption (W)},
        legend image code/.code={
            \draw[#1, bar width=6pt, yshift=-0.3em] plot coordinates {(0cm,0.8em)};
        }
    ]
            
        \nextgroupplot [
            width=0.34\textwidth,
		    xticklabels from table={data/energy_efficiency.txt}{Name}
        ]
        
        \addplot [thick, pattern=vertical lines, pattern color=red] table [x=id, y=static] {data/energy_efficiency.txt};
        \label{plt:power_static}
        \addplot [thick, pattern=horizontal lines, pattern color=blue] table [x=id, y=dynamic] {data/energy_efficiency.txt}
        node [pos=0.2, rotate=90, anchor=west] {4.43$\times$}
        node [pos=0.4, rotate=90, anchor=west] {5.04$\times$}
        node [pos=0.6, rotate=90, anchor=west] {5.85$\times$}
        node [pos=0.8, rotate=90, anchor=west] {6.66$\times$}
        node [pos=1, rotate=90, anchor=west] {6.37$\times$};
        \label{plt:power_dynamic}
        
        \node [text width=1em, anchor=north] at (axis description cs:0.5, 1)
        {\subcaption{\label{plot:power_unrolled}}};
            
        \nextgroupplot [
            width=0.56\textwidth,
            ymax=8,
            xticklabels from table={data/energy_efficiency_tiled.txt}{Name},
        ]
        
        \addplot [thick, pattern=vertical lines, pattern color=red] table [x=id, y=static] {data/energy_efficiency_tiled.txt};
        \addplot [thick, pattern=horizontal lines, pattern color=blue] table [x=id, y=dynamic] {data/energy_efficiency_tiled.txt}
        node [pos=0.111, rotate=90, anchor=west] {1.27$\times$}
        node [pos=0.222, rotate=90, anchor=west] {1.29$\times$}
        node [pos=0.333, rotate=90, anchor=west] {1.48$\times$}
        node [pos=0.444, rotate=90, anchor=west] {1.57$\times$}
        node [pos=0.556, rotate=90, anchor=west] {1.49$\times$}
        node [pos=0.667, rotate=90, anchor=west] {1.10$\times$}
        node [pos=0.778, rotate=90, anchor=west] {1.04$\times$}
        node [pos=0.889, rotate=90, anchor=west] {1.04$\times$}
        node [pos=1, rotate=90, anchor=west] {1.08$\times$};
        
        \node [text width=1em, anchor=north] at (axis description cs:0.5, 1)
        {\subcaption{\label{plot:power_tiled}}};
        
    \end{groupplot}

\end{tikzpicture}
            	\caption{
            	    Implementation power consumption estimates, broken into static~(\ref{plt:power_static}) and dynamic~(\ref{plt:power_dynamic}) components, for the same CNV implementations shown in Fig.~\ref{plot:breakdown}.
            	    In \subref{plot:power_unrolled}, designs were unrolled, while those in \subref{plot:power_tiled} were tiled with $T_\text{i} = T_\text{o} = 8$.
                	Annotations show decreases \emph{vs} ReBNet~\cite{BNN_CNN_REBNET_FCCM}.
                }
                \label{plot:power}
            \end{figure*}
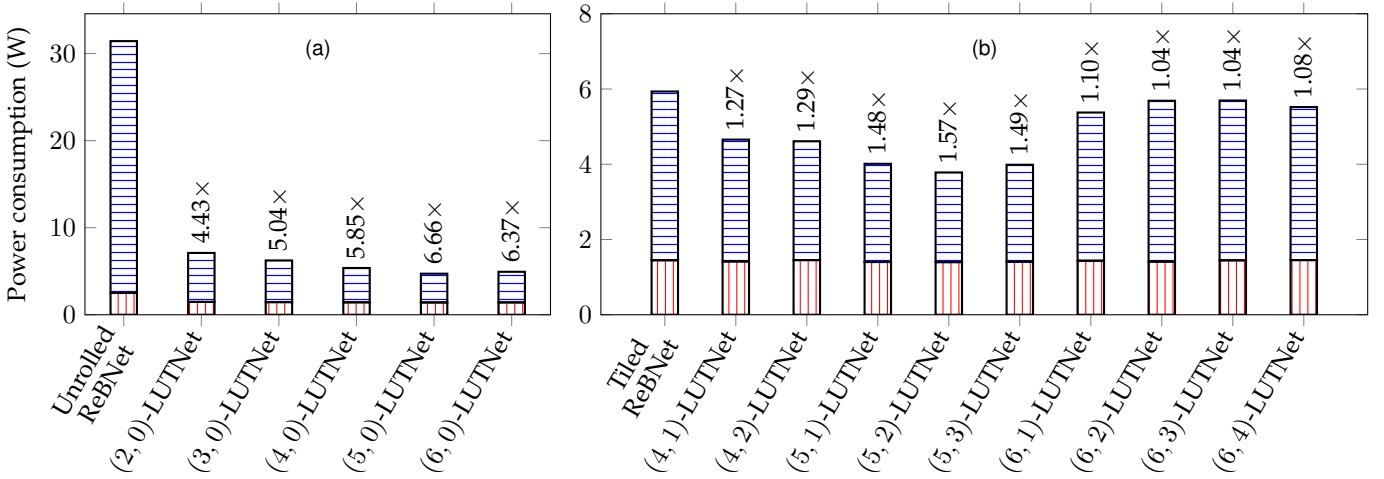
			
			While Fig.~\ref{plot:robustness_ti8} (fixed $T_\text{i}$) and \subref{plot:robustness_to8} (fixed $T_\text{o}$) show similar trends, a slight preference towards increasing $T_\text{o}$ is evident.
			This is reflected in the more heavily pruned cases, where both $\left(T_\text{i},T_\text{o}\right) = \left(8,4\right)$ and $\left(16,8\right)$ jump up in error sooner than their respective counterparts, $\left(4,8\right)$ and $\left(8,16\right)$.
			When tiling across only inputs, weights are shared within units computing each output.
			With output-wise tiling, however, they are instead shared across units calculating different outputs; there is no intra-unit weight sharing in this case.
			When pruning to some predetermined level, the likelihood of entire output channels being thinned is therefore higher in the case of input-only tiling, leading to greater accuracy degradation.
			The same argument holds for cases where tiling across both inputs and outputs is applied, but with a higher degree of the former over the latter.
			Thus, when one is faced with a choice between increasing either $T_\text{i}$ or $T_\text{o}$, we suggest that prioritization should be given to $T_\text{o}$.
		
		\subsection{Energy Efficiency}
		\label{sec:eval_energy}
			
			We estimated LUTNet's energy efficiency using the Xilinx Power Analyzer (XPA) tool with default settings: vectorless mode and 12.5\% primary input switching probability.
			These are commonly used for early power estimation in industry~\cite{Davis_FPL17}.
			The resultant power estimates, for the same implementations captured in Fig.~\ref{plot:breakdown}, are shown in Fig.~\ref{plot:power}.
            All were obtained post-placement and \-/routing.
            Power consumption is equivalent to energy efficiency here since all implementations presented side-by-side have identical throughput.
            While vectorless power estimates are not particularly accurate---typically around $\pm$10--20\% from measured values~\cite{KAPOW}---they are sufficiently so for our purposes.
				
            
            Since dynamic power consumption is directly related to area occupancy, Figs.~\ref{plot:breakdown_unrolled} and \ref{plot:power_unrolled} show similar trends.
            Most of the fully unrolled networks' area consumption is attributable to popcount adder trees, whose carry chains are dominant with respect to switching activity.
            Popcount branch pruning shortens the chains, more than proportionately lowering their switching rates and thereby causing the large dynamic power drop.
            The reduction in static power between the unrolled ReBNet and LUTNet implementations can also be linked to area, although indirectly.
            Between Pruned ReBNet and $\left(2,0\right)$-LUTNet there was a drop in estimated junction temperature from 60.1\textdegree~to 31.3\textdegree, leading to reduced leakage current and therefore static power draw.
            Such temperature decreases are also useful since they limit aging, thereby increasing device lifetime~\cite{DEGRADATION}.
            Overall, we can conclude that unrolled LUTNet implementations' significant area reductions result in even greater energy efficiency improvements over their ReBNet counterparts.
        
            In the same style as Fig.~\ref{plot:power_unrolled}, Fig.~\ref{plot:power_tiled} shows power consumption estimates for the tiled implementations whose area breakdowns were reported in Fig.~\ref{plot:breakdown_tiled}.
            Comparing between Fig.~\ref{plot:power_unrolled} and \subref{plot:power_tiled}, we can see large jumps in the ratio of static \emph{vs} dynamic power for the latter over the former.
            This is a direct consequence of the tiled designs' greatly reduced resource requirements.
            Dynamic power reductions over ReBNet were modest compared to those for the unrolled implementations, which tended to be large: up to 8.76$\times$.
            Despite this, however, we still achieved energy efficiency improvements over ReBNet in all cases.
            Notice also that the total power consumptions of tiled LUTNet implementations are similar to those of the unrolled alternatives.
            For example, unrolled $\left(5,0\right)$-LUTNet is estimated to consume 4.72~W, while $\left(5,1\right)$-LUTNet with $T_\text{i} = T_\text{o} = 8$ consumes 4.01~W.
            Although unrolled implementations occupy more area than their tiled counterparts, the former's greatly increased resilience to pruning---coupled with throughput-bottlenecking by other network layers---makes them similarly power-hungry, despite being faster.
		
		\subsection{Throughput Maximization}
		\label{sec:eval_throughput}
			
			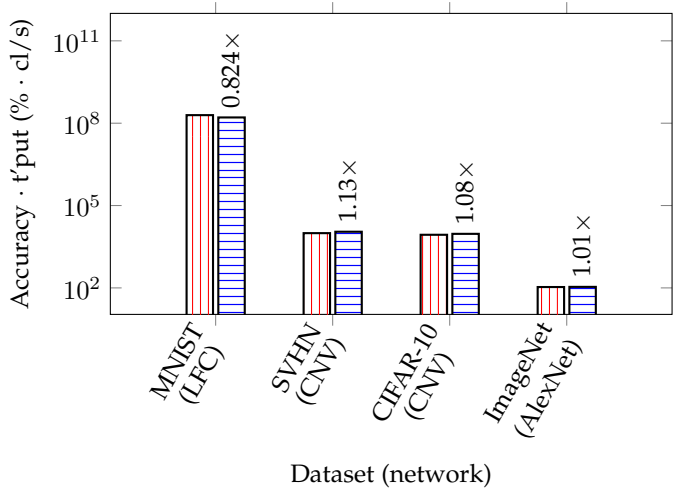
\begin{figure}
                \centering
                \begin{tikzpicture}
    
    \begin{axis}[ybar,
        scale only axis,
		width=0.84\columnwidth,
		height=0.22\textwidth,
		ymode=log,
        ymin=0,
        ymax=1000000000000,
        ylabel near ticks,
		ylabel={Accuracy $\cdot$ t'put (\% $\cdot$ cl/s)},
		xtick={1, 2, 3, 4},
        xticklabels={\shortstack{MNIST\\(LFC)},\shortstack{SVHN\\(CNV)},\shortstack{CIFAR-10\\(CNV)},\shortstack{ImageNet\\(AlexNet)}},
        enlarge x limits=0.3,
        x tick label style={rotate=60, anchor=east},
        xlabel near ticks,
        xlabel={Dataset (network)},
        legend image code/.code={
            \draw[#1, bar width=6pt, yshift=-0.3em] plot coordinates {(0cm,0.8em)};
        }
    ]
        \addplot [thick, pattern=vertical lines, pattern color=red] coordinates  {(1, 196680000) (2, 9879.048219) (3, 8616.840048) (4, 107.562)};
        \label{plt:area_comp_tiled_rebnet_wup};
        \addplot [thick, pattern=horizontal lines, pattern color=blue] coordinates  {(1, 162031792) (2, 11121.88046) (3, 9319.160409) (4, 108.9905655)}
        node [pos=0, xshift=0.6*\pgfkeysvalueof{/pgf/bar width}, rotate=90, anchor=west] {0.824$\times$}
        node [pos=0.333, xshift=0.6*\pgfkeysvalueof{/pgf/bar width}, rotate=90, anchor=west] {1.13$\times$}
        node [pos=0.667, xshift=0.6*\pgfkeysvalueof{/pgf/bar width}, rotate=90, anchor=west] {1.08$\times$}
        node [pos=1, xshift=0.6*\pgfkeysvalueof{/pgf/bar width}, rotate=90, anchor=west] {1.01$\times$};
        \label{plt:area_comp_tiled_4lutnet};
    \end{axis}

\end{tikzpicture}
            	\caption{
            	    Accuracy-throughput product for ReBNet~\cite{BNN_CNN_REBNET_FCCM}~(\ref{plt:area_comp_tiled_rebnet_wup}) and $\left(5,1\right)$-LUTNet~(\ref{plt:area_comp_tiled_4lutnet}) implementations with various models and datasets.
            	    For each design, layer-wise pruning thresholds $\theta$ and tiling factors $T_\text{i}$ and $T_\text{o}$ were chosen to attempt to maximize accuracy per unit throughput by saturating our target FPGA's resources while keeping throughput balanced across layers.
            	    Annotations show increases.
                }
                \label{plot:throughput}
            \end{figure}
            
            \begin{figure*}
                \centering
    


\begin{tikzpicture}
    
    \begin{groupplot}[
        ybar stacked,
        group style={
            columns=2,
            ylabels at=edge left,
            yticklabels at=edge left,
            horizontal sep=1em
        },
        scale only axis,
        height=0.22\textwidth,
        ymin=0,
        ymax=120,
        xtick=data,
        x tick label style={rotate=60, anchor=east},
        xtick align=outside,
        ylabel near ticks,
        ylabel={Training time per epoch (s)},
        legend image code/.code={
            \draw[#1, bar width=6pt, yshift=-0.3em] plot coordinates {(0cm,0.8em)};
        }
    ]
            
        \nextgroupplot [
            width=0.34\textwidth,
		    xticklabels from table={data/training_speed.txt}{Name}
        ]
        
        \addplot [thick] table [x=id, y={training_time}] {data/training_speed.txt}
        node [pos=0.20, rotate=90, anchor=west] {1.00$\times$}
        node [pos=0.40, rotate=90, anchor=west] {1.03$\times$}
        node [pos=0.60, rotate=90, anchor=west] {1.03$\times$}
        node [pos=0.80, rotate=90, anchor=west] {1.67$\times$}
        node [pos=1, rotate=90, anchor=west] {2.24$\times$};
        
        \node [text width=1em, anchor=north] at (axis description cs:0.5, 1)
        {\subcaption{\label{plot:training_time_unrolled}}};
            
        \nextgroupplot [
            width=0.57\textwidth,
            xticklabels from table={data/training_speed_tiled.txt}{Name}
        ]
        
        \addplot [thick] table [x=id, y={training_time}] {data/training_speed_tiled.txt}
        node [pos=0.111, rotate=90, anchor=west] {1.03$\times$}
        node [pos=0.222, rotate=90, anchor=west] {1.09$\times$}
        node [pos=0.333, rotate=90, anchor=west] {1.27$\times$}
        node [pos=0.444, rotate=90, anchor=west] {1.58$\times$}
        node [pos=0.556, rotate=90, anchor=west] {1.48$\times$}
        node [pos=0.667, rotate=90, anchor=west] {2.30$\times$}
        node [pos=0.778, rotate=90, anchor=west] {1.91$\times$}
        node [pos=0.889, rotate=90, anchor=west] {2.27$\times$}
        node [pos=1, rotate=90, anchor=west] {2.58$\times$};
        
        \node [text width=1em, anchor=north] at (axis description cs:0.5, 1)
        {\subcaption{\label{plot:training_time_tiled}}};
        
    \end{groupplot}

\end{tikzpicture}
            	\caption{
            	    Training time per epoch for the same CNV implementations shown in Figs.~\ref{plot:breakdown} and \ref{plot:power}.
                	Annotations show increases \emph{vs} ReBNet~\cite{BNN_CNN_REBNET_FCCM}.
                }
                \label{plot:training}
            \end{figure*}
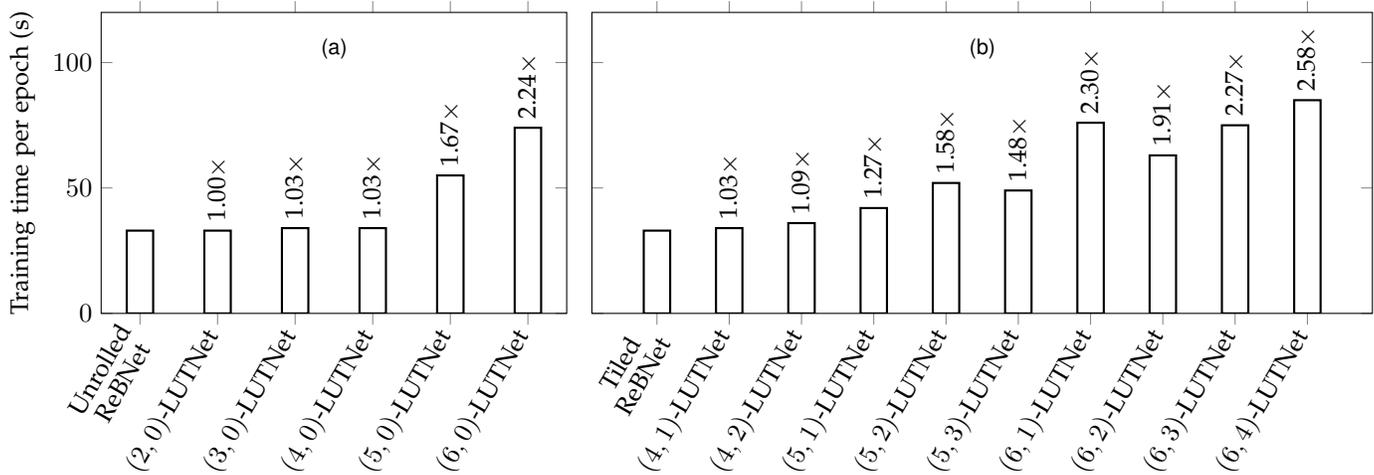
			
			In an effort to demonstrate the potential of LUTNet for whole-network realization, we created implementations for each of the network-dataset pairs detailed in Table~\ref{tab:model_info}, using our proposed architecture for all convolutional and fully connected layers.
			Our aim was to, within the area constraints imposed by our target FPGA, maximize both throughput (in classifications per second, cl/s) and top-1 test accuracy.
			We combined these into a single metric, accuracy-throughput product, for comparison to the ReBNet implementations that we constructed in the same way.
			
			The experiments reported in Fig.~\ref{AREA_LUT_TRADEOFF_SEARCH_K4} revealed that, of the feasible combinations of $\left(K,P\right)$ for tiled LUTNet, $\left(5,1\right)$ behaved the most favorably in terms of area \emph{vs} accuracy when pruned.
			For completeness, we performed the same experiments for $K = 2$ (not shown in Fig.~\ref{AREA_LUT_TRADEOFF_SEARCH_K4}) and confirmed this to be the case.
			Thus, for this section, we used $\left(K,P\right) = \left(5,1\right)$ throughout.
			For each layer within each benchmark network, we hand-tuned pruning threshold $\theta$ and tiling factors $T_\text{i}$ and $T_\text{o}$ to maintain high accuracy and keep throughput balanced across layers while making as much use of the available resources as possible.
			We present the results of these experiments in Fig.~\ref{plot:throughput}, which shows modest improvements over ReBNet implementations for all models except for LFC classifying MNIST, in common with the result for the same network-dataset pair seen in Fig.~\ref{plot:area_compression}.
			
			
			We stress that these results are not optimal.
			Due to the sheer size of our design space and the lengthy training times associated with whole-network implementation, it was not practicable for us to locate the best design points possible.
			Based on our findings from the parameter tuning of limited numbers of layers, we made educated guesses of sensible choices for $K$, $P$, $\theta$, and $T_\text{i}$ \emph{vs} $T_\text{o}$ ratios.
			With time and compute power dedicated to design-space exploration, greater performance gains could doubtless be achieved.
		
		\subsection{Training Time}
			
			We finally sought to quantify the primary cost of LUTNet's deployment: that of training time.
			In Fig.~\ref{plot:training}, we present the whole-network training times of the implementations used to generate the results in Figs.~\ref{plot:breakdown} and \ref{plot:power}.
				
			
			Each of $\left(K,0\right)$-LUTNet's inference $K$-LUTs consists of 2\textsuperscript{$K$} parameters: 2$\times$ more than that for $\left(K-1,0\right)$-LUTNet.
            Consequently, the number of training operations required per epoch increases exponentially with $K$.
            This does not necessarily translate to exponentially increasing training times over XNOR-based BNNs, however, since, as pointed out by Jouppi \emph{et al.}, the majority of DNN training accelerators' speed is bounded by memory bandwidth, not compute power~\cite{FXP_CNN_TPU}.
            This is evident from Fig.~\ref{plot:training_time_unrolled}, which shows the per-epoch training times of ReBNet and LUTNet implementations for CNV with CIFAR-10.
            Implementations from ReBNet to $\left(4,0\right)$-LUTNet all have approximately the same training rate, despite the number of parameters increasing by up to 16$\times$.
            The training time did not increase because, for all of these implementations, progress was bottlenecked by high-precision activation transfer to and from GPU RAM.
            Increases of significance were seen for $\left(5,0\right)$-LUTNet and beyond, for which the number of multiply-accumulate operations performed per activation transferred rose enough for the former to dominate.
            
            Tiled implementations take longer to train than those with the same layers unrolled, as reflected in Fig.~\ref{plot:training_time_tiled}.
            This is unsurprising due to tiling's effect on the number of trainable parameters within the network, the most significant component of which increases linearly with $P$, $T_\text{i}$, and $T_\text{o}$, as discussed in Section~\ref{sec:eval_area_tiled}.
            It is interesting to note that the increases in training time with $P$ are not as significant as the parameter growth rate with $P T_\text{i} T_\text{o}$ might suggest.
            In every epoch, TensorFlow stores a new copy of input activations in RAM for each LUT's $K - P$ input connections.
            As $P$ grows, $K - P$ reduces for the same $K$, thus increasing the number of operations to perform but decreasing the number of these expensive memory copies.
            We see that the latter thus dampens the slowdowns brought about by the former.
            
            Recall that all of the designs featured in Fig.~\ref{plot:training} only have a single layer implemented using LUTNet operators.
            Slowdowns in training increase significantly with, in particular, whole-network LUTNet deployment.
            For example, while a CNV implementation with only the largest convolutional layer implemented using the unrolled $\left(5,0\right)$-LUTNet architecture takes 1.67$\times$ longer to train than a fully ReBNet equivalent, this factor increases to 15.8 when the same network is wholly implemented using $(5,1)$-LUTNet inference operators with $T_\text{i} = T_\text{o} = 8$.
            We do not consider this to be of significant concern, however: training times of around a day are not of dissimilar duration to compilation times for large FPGA designs, which will be needed as well, and are far shorter than their typical development cycles.

	\section{Limitations}
	
		Figs.~\ref{plot:AREA_LUT_TRADEOFF} and \ref{AREA_LUT_TRADEOFF_SEARCH_K4} show that while our expansion to 2-LUTs results in significant logic density gains over XNORs, returns for movement to $K$-LUTs for $K > 2$ are diminishing.
        We suspect that this is due to our current restriction on the form of the function $g_m$ in \eqref{eq:synapse_tm_lutnet}, \emph{i.e.} $\left\{-1,1\right\}^{K-P} \times \left\{-1,1\right\}^P \to \left\{-1,1\right\}$ rather than $\left\{-1,1\right\}^{K-P} \times \left\{-1,1\right\}^P \to \mathbb{N}$.
        This makes \eqref{eqn:retrain_function} insoluble when $\hat{c}_{\boldsymbol{d}}$ and $\hat{\tilde{p}}^{\left(m,t\right)}$ are restricted to binary values.
        We could overcome this, and potentially make even more efficient use of the underlying FPGA fabric, by learning the popcount circuitry along with our XNOR substitutes, replacing the summation as well as $w_n x_n$ in \eqref{eq:synapse_normal}.
        
		While the introduction of nonlinearity significantly increases the expressiveness of each inference operator, the experiments reported in Section~\ref{sec:eval_area} revealed that $\left(6,P\right)$-LUTNet showed early signs of overfitting.
        In the future, we will explore methods of throttling expressiveness during training guided by losses, \emph{e.g.} switching to higher or lower values of $K$ and $P$ where appropriate.
        
        Post-logic expansion analysis of our implementations showed that the majority of functions performed by our inference $K$-LUTs remained XNOR.
        Given the freedom afforded by our training regime, this was somewhat surprising.
        We suspect that our current network initialization strategy, in which all operations begin as XNOR, causes this by biasing gradient descent.
        In the hope of increasing expressiveness to facilitate more aggressive pruning, we will explore alternative initializations to combat this effect.
        
        Finally, LUTNet's software does not currently skip zeros during training.
        As networks increase in size, GPU RAM will be increasingly inefficiently used, resulting in unnecessarily long training times.
        A future revision will therefore incorporate sparse matrix multiplication.

    \section{Conclusion}
	
        In this article, we introduced LUTNet: the first DNN architecture featuring $K$-LUTs as inference operators specifically designed to suit FPGA implementation.
        Our novel training approach results in the construction of $K$-LUT-based networks robust to high levels of pruning with little or no accuracy degradation, enabling the achievement of significantly higher area and energy efficiency than BNNs'.
        
        We comprehensively evaluated both unrolled and tiled versions of the LUTNet architecture.
        For the former, our experiments with 4-LUT-based inference operators revealed that FPGA implementations following our proposals achieved a mean area reduction of 1.81$\times$ \emph{vs} the state-of-the-art BNN architecture with unrolling and pruning.
        These designs targeted a range of standard DNN models and datasets, required approximately the same training time, and reached accuracy bounded within $\pm$0.300~pp in all cases.
        Due to their efficient use of soft logic, unrolled LUTNet implementations can exhibit energy efficiency up to 6.66$\times$ greater than reported by the authors of related prior works.
        Thanks to its parameter hardening, our unrolled architecture also requires no use of block RAM: a common bottleneck for FPGA-deployed DNNs.
        
        While unrolled LUTNet implementations have many advantages over their tiled counterparts---particularly higher throughput and compressibility---they typically consume large numbers of resources.
        When tiled, LUTNet designs can achieve similarly high accuracy while occupying much less area.
        For example, we found that an $8 \times 8$-tiled and 64.6\%-pruned $\left(5,1\right)$-LUTNet CNV implementation was $2.78\times$ smaller and $1.18\times$ less power-hungry than an unrolled and 91.1\%-pruned $\left(4,0\right)$-LUTNet equivalent, with both delivering comparable CIFAR-10 classification accuracy.
        The choice of whether or not to tile, and to what level if so, will depend on the amount of LUT and RAM resources one can afford to use.
        With increased tiling, throughput and area efficiency can be sacrificed to achieve smaller, lower-power designs.
        
        The authors of existing works on low-precision DNN inference seem to have assumed that their forward-propagation functions must be good approximations of the linear dot product.
        With LUTNet, we argue for a tangential approach: through the embracement of nonlinearity, one can do more with less by unlocking the full potential of the LUT.
		
    \bibliographystyle{IEEEtran}
    \bibliography{pynq_cnn_bibliography}

\begin{thebibliography}{10}
\providecommand{\url}[1]{#1}
\csname url@samestyle\endcsname
\providecommand{\newblock}{\relax}
\providecommand{\bibinfo}[2]{#2}
\providecommand{\BIBentrySTDinterwordspacing}{\spaceskip=0pt\relax}
\providecommand{\BIBentryALTinterwordstretchfactor}{4}
\providecommand{\BIBentryALTinterwordspacing}{\spaceskip=\fontdimen2\font plus
\BIBentryALTinterwordstretchfactor\fontdimen3\font minus
  \fontdimen4\font\relax}
\providecommand{\BIBforeignlanguage}[2]{{%
\expandafter\ifx\csname l@#1\endcsname\relax
\typeout{** WARNING: IEEEtran.bst: No hyphenation pattern has been}%
\typeout{** loaded for the language `#1'. Using the pattern for}%
\typeout{** the default language instead.}%
\else
\language=\csname l@#1\endcsname
\fi
#2}}
\providecommand{\BIBdecl}{\relax}
\BIBdecl

\bibitem{Parhi}
K.~K. Parhi, \emph{{VLSI} Digital Signal Processing Systems: Design and
  Implementation}.\hskip 1em plus 0.5em minus 0.4em\relax Wiley, 1999.

\bibitem{BNN_CNN_FINN}
Y.~Umuroglu, N.~J. Fraser, G.~Gambardella, M.~Blott, P.~H.~W. Leong, M.~Jahre,
  and K.~Vissers, ``{FINN}: A framework for fast, scalable binarized neural
  network inference,'' in \emph{ACM/SIGDA International Symposium on
  Field-Programmable Gate Arrays}, 2017.

\bibitem{ALEXNET}
A.~Krizhevsky, I.~Sutskever, and G.~E. Hinton, ``{ImageNet} classification with
  deep convolutional neural networks,'' in \emph{Conference on Neural
  Information Processing Systems}, 2012.

\bibitem{ASAP}
R.~Zhao, S.~Liu, H.-C. Ng, E.~Wang, J.~J. Davis, X.~Niu, X.~Wang, H.~Shi, G.~A.
  Constantinides, P.~Y.~K. Cheung, and W.~Luk, ``Hardware compilation of deep
  neural networks: An overview,'' in \emph{International Conference on
  Application-specific Systems, Architectures and Processors}, 2018.

\bibitem{CSUR}
E.~Wang, J.~J. Davis, R.~Zhao, H.-C. Ng, X.~Niu, W.~Luk, P.~Y.~K. Cheung, and
  G.~A. Constantinides, ``Deep neural network approximation for custom
  hardware: Where we've been, where we're going,'' \emph{ACM Computing
  Surveys}, vol.~52, no.~2, 2019.

\bibitem{STYLIANOS_TOOLFLOWS}
S.~I. Venieris, A.~Kouris, and C.-S. Bouganis, ``Toolflows for mapping
  convolutional neural networks on {FPGAs}: A survey and future directions,''
  \emph{ACM Computing Surveys}, vol.~51, no.~3, 2018.

\bibitem{SURV_FPGA_BASED_NEURAL_NETWORK_ACC}
K.~Guo, S.~Zeng, J.~Yu, Y.~Wang, and H.~Yang, ``A survey of {FPGA} based neural
  network accelerator,'' \emph{ACM Transactions on Reconfigurable Technology
  and Systems}, vol.~9, no.~4, 2017.

\bibitem{CIFAR10}
A.~Krizhevsky, ``Learning multiple layers of features from tiny images,''
  Master's thesis, University of Toronto, 2009.

\bibitem{IMAGENET}
J.~Deng, W.~Dong, R.~Socher, J.~Li, K.~Li, and F.~Li, ``{ImageNet}: A
  large-scale hierarchical image database,'' in \emph{IEEE Conference on
  Computer Vision and Pattern Recognition}, 2009.

\bibitem{BNN_CNN_REBNET_FCCM}
M.~Ghasemzadeh, M.~Samragh, and F.~Koushanfar, ``{ReBNet}: Residual binarized
  neural network,'' in \emph{IEEE International Symposium on Field-Programmable
  Custom Computing Machines}, 2018.

\bibitem{LUTNET}
E.~Wang, J.~J. Davis, P.~Y.~K. Cheung, and G.~A. Constantinides, ``{LUTNet}:
  Rethinking inference in {FPGA} soft logic,'' in \emph{IEEE International
  Symposium on Field-Programmable Custom Computing Machines}, 2019.

\bibitem{BNN_CNN_BinaryConnect}
M.~Courbariaux, Y.~Bengio, and J.-P. David, ``{BinaryConnect}: Training deep
  neural networks with binary weights during propagations,'' in
  \emph{Conference on Neural Information Processing Systems}, 2015.

\bibitem{BNN_CNN_BinaryNet}
M.~Courbariaux and Y.~Bengio, ``{BinaryNet}: Training deep neural networks with
  weights and activations constrained to +1 or -1,'' \emph{arXiv preprint
  arXiv:1602.02830}, 2016.

\bibitem{AARON_ZHAO_FPT}
Y.~Zhao, X.~Gao, X.~Guo, J.~Liu, E.~Wang, R.~Mullins, P.~Y.~K. Cheung,
  G.~Constantinides, and C.-Z. Xu, ``Automatic generation of multi-precision
  multi-arithmetic {CNN} accelerators for {FPGAs},'' in \emph{International
  Conference on Field-Programmable Technology}, 2019.

\bibitem{AARON_ZHAO_NIPS}
Y.~Zhao, X.~Gao, D.~Bates, R.~Mullins, and C.-Z. Xu, ``Focused quantization for
  sparse {CNNs},'' in \emph{Advances in Neural Information Processing Systems},
  2019.

\bibitem{BNN_CNN_BINARY_CONSTRIANED_TRAINING}
W.~Tang, G.~Hua, and L.~Wang, ``How to train a compact binary neural network
  with high accuracy?'' in \emph{Association for the Advancement of Artificial
  Intelligence}, 2017.

\bibitem{PRU_CNN_TRAIN_PRUNE_RETRAIN}
S.~Han, J.~Pool, J.~Tran, and W.~J. Dally, ``Learning both weights and
  connections for efficient neural network,'' in \emph{Conference on Neural
  Information Processing Systems}, 2015.

\bibitem{BNN_CNN_XNOR-Net}
M.~Rastegari, V.~Ordonez, J.~Redmon, and A.~Farhadi, ``{XNOR-Net}: Imagenet
  classification using binary convolutional neural networks,'' in
  \emph{European Conference on Computer Vision}, 2016.

\bibitem{BNN_CNN_ABC-Net}
X.~Lin, C.~Zhao, and W.~Pan, ``Towards accurate binary convolutional neural
  network,'' in \emph{Conference on Neural Information Processing Systems},
  2017.

\bibitem{TNN_CNN_TWN}
F.~Li and B.~Liu, ``Ternary weight networks,'' in \emph{Conference on Neural
  Information Processing Systems}, 2016.

\bibitem{TNN_CNN_TTQ}
C.~Zhu, S.~Han, H.~Mao, and W.~J. Dally, ``Trained ternary quantization,'' in
  \emph{International Conference on Learning Representations}, 2017.

\bibitem{TNN_CNN_BENGIO}
Z.~Lin, M.~Courbariaux, R.~Memisevic, and Y.~Bengio, ``Neural networks with few
  multiplications,'' in \emph{International Conference on Learning
  Representations}, 2015.

\bibitem{PRU_CNN_STRUCTURED_SPARSITY}
W.~Wen, C.~Wu, Y.~Wang, Y.~Chen, and H.~Li, ``Learning structured sparsity in
  deep neural networks,'' in \emph{Conference on Neural Information Processing
  Systems}, 2016.

\bibitem{NR_CNN_FXP_OPTIM_LOOP_JASON_CONG}
C.~Zhang, P.~Li, G.~Sun, Y.~Guan, B.~Xiao, and J.~Cong, ``Optimizing
  {FPGA}-based accelerator design for deep convolutional neural networks,'' in
  \emph{ACM/SIGDA International Symposium on Field-Programmable Gate Arrays},
  2015.

\bibitem{NR_CNN_FXP_OPTIM_LOOP}
Y.~Ma, Y.~Cao, S.~Vrudhula, and J.-S. Seo, ``Optimizing loop operation and
  dataflow in fpga acceleration of deep convolutional neural networks,'' in
  \emph{ACM/SIGDA International Symposium on Field-Programmable Gate Arrays},
  2017.

\bibitem{BETZ_FPGA19}
A.~Boutros, M.~Eldafrawy, S.~Yazdanshenas, and V.~Betz, ``Math doesn't have to
  be hard: Logic block architectures to enhance low-precision
  multiply-accumulate on {FPGAs},'' in \emph{ACM/SIGDA International Symposium
  on Field-Programmable Gate Arrays}, 2019.

\bibitem{PIRDSP_FCCM19}
S.~Rasoulinezhad, H.~Zhou, L.~Wang, and P.~H.~W. Leong, ``{PIR-DSP}: An {FPGA}
  {DSP} block architecture for multi-precision deep neural networks,'' in
  \emph{International Symposium on Field-Programmable Custom Computing
  Machines}, 2019.

\bibitem{ZHANG_RETHINKING_GENERALISATION}
C.~Zhang, S.~Bengio, M.~Hardt, B.~Recht, and O.~Vinyals, ``Understanding deep
  learning requires rethinking generalization,'' in \emph{International
  Conference on Learning Representations}, 2017.

\bibitem{CHATTERJEE_MEMORIZATION}
S.~Chatterjee, ``Learning and memorization,'' in \emph{International Conference
  on Machine Learning}, 2018.

\bibitem{NR_CNN_FXP_FPGACONVNET}
S.~I. Venieris and C.-S. Bouganis, ``{fpgaConvNet}: A framework for mapping
  convolutional neural networks on {FPGAs},'' in \emph{IEEE International
  Symposium on Field-Programmable Custom Computing Machines}, 2016.

\bibitem{MNIST}
Y.~LeCun, L.~Bottou, Y.~Bengio, and P.~Haffner, ``Gradient-based learning
  applied to document recognition,'' \emph{{Proceedings of the IEEE}}, 1998.

\bibitem{Davis_FPL17}
J.~J. Davis, J.~M. Levine, E.~A. Stott, E.~Hung, P.~Y.~K. Cheung, and G.~A.
  Constantinides, ``{STRIPE}: Signal selection for runtime power estimation,''
  in \emph{International Conference on Field-Programmable Logic and
  Applications}, 2017.

\bibitem{KAPOW}
J.~J. Davis, E.~Hung, J.~M. Levine, E.~A. Stott, P.~Y.~K. Cheung, and G.~A.
  Constantinides, ``{KAPow}: High-accuracy, low-overhead online per-module
  power estimation for {FPGA} designs,'' \emph{ACM Transactions on
  Reconfigurable Technology and Systems}, vol.~11, no.~1, 2018.

\bibitem{DEGRADATION}
E.~Stott, J.~S.~J. Wong, and P.~Y.~K. Cheung, ``Degradation analysis and
  mitigation in {FPGAs},'' in \emph{International Conference on
  Field-Programmable Logic and Applications}, 2010.

\bibitem{FXP_CNN_TPU}
N.~P. Jouppi, C.~Young, N.~Patil, D.~Patterson, G.~Agrawal, R.~Bajwa, S.~Bates,
  S.~Bhatia, N.~Boden, and A.~Borchers, ``In-datacenter performance analysis of
  a {Tensor Processing Unit},'' in \emph{International Symposium on Computer
  Architecture}, 2017.

\end{thebibliography}
    
    \begin{IEEEbiography}[{\includegraphics[width=1in,height=1.25in,clip,keepaspectratio]{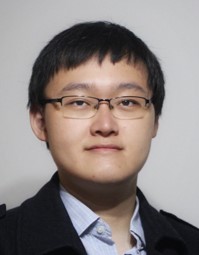}}]{Erwei Wang}
        (S'18)
		is a PhD student in the Department of Electrical and Electronic Engineering's Circuits and Systems group at Imperial College London.
		His research interests include deep neural networks, computer vision systems, and high-performance computing architectures, with an emphasis on improving speed and energy efficiency for custom hardware implementation.
		He is a Student Member of the IEEE.
	\end{IEEEbiography}
    
    \begin{IEEEbiography}[{\includegraphics[width=1in,height=1.25in,clip,keepaspectratio]{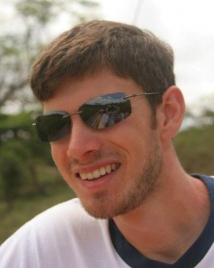}}]{James J. Davis}
		(S'08--M'16)
		is a Research Fellow in the Department of Electrical and Electronic Engineering's Circuits and Systems group at Imperial College London.
		He received a PhD in Electrical and Electronic Engineering from Imperial College London in 2016.
		His research is focussed on the exploitation of FPGA features for cutting-edge applications, driving up performance, energy efficiency, and reliability.
		Dr Davis serves on the technical program committees of the four top-tier reconfigurable computing conferences (FPGA, FCCM, FPL, and FPT) and is a multi-best paper award recipient.
		He is a Member of the IEEE and the ACM.
	\end{IEEEbiography}
	
	\begin{IEEEbiography}[{\includegraphics[width=1in,height=1.25in,clip,keepaspectratio]{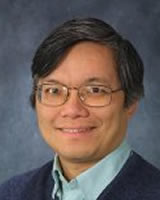}}]{Peter Y. K. Cheung}
		(M'85--SM'04)
		is Professor of Digital Systems in the Department of Electrical and Electronic Engineering at Imperial College London.
		His research interests include VLSI architectures for signal processing, asynchronous systems, reconfigurable computing, and architectural synthesis.
		Prof. Cheung received a DSc from Imperial College London.
		He is a Senior Member of the IEEE and Fellow of the IET.
	\end{IEEEbiography}
	
	\begin{IEEEbiography}[{\includegraphics[width=1in,height=1.25in,clip,keepaspectratio]{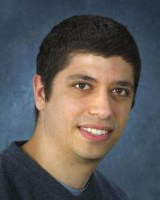}}]{George A. Constantinides}
		(S'96--M'01--SM'08)
		received the PhD degree from Imperial College London in 2001.
		Since 2002, he has been with the faculty at Imperial College London, where he is currently Professor of Digital Computation and Head of the Circuits and Systems research group.
		He was the general chair of the ACM International Symposium on Field-Programmable Gate Arrays in 2015.
		He serves on several program committees and has published over 200 research papers in peer-refereed journals and international conferences.
		Prof. Constantinides is a Senior Member of the IEEE and a Fellow of the BCS.
	\end{IEEEbiography}

\end{document}